\documentclass[lettersize,journal]{IEEEtran}
\usepackage{amsmath,amsfonts}
\usepackage{algorithmic}
\usepackage{algorithm}
\usepackage{array}
\usepackage{textcomp}
\usepackage{stfloats}
\usepackage{url}
\usepackage{verbatim}
\usepackage{graphicx}
\usepackage{subfigure}
\usepackage[compress]{cite} 
\usepackage{float}
\usepackage{booktabs}
\usepackage{multirow}
\usepackage{makecell}
\usepackage{tikz,xcolor}
\usepackage[colorlinks,linkcolor=blue]{hyperref}
\hypersetup{hidelinks,
	colorlinks=true,
	allcolors=black,
	pdfstartview=Fit,
	breaklinks=true}
\definecolor{lime}{HTML}{A6CE39}
\DeclareRobustCommand{\orcidicon}{
\begin{tikzpicture}
\draw[lime, fill=lime] (0,0)
circle[radius=0.16]
node[white]{{\fontfamily{qag}\selectfont \tiny \.{I}D}};
\end{tikzpicture}
\hspace{-2mm}
}
\foreach \x in {Ronghao,Huhaifeng,Sijie,A, ..., Z}{%
\expandafter\xdef\csname orcid\x\endcsname{\noexpand\href{https://orcid.org/\csname orcidauthor\x\endcsname}{\noexpand\orcidicon}}
}

\begin{document}
\title{End-to-end Semantic-centric Video-based Multimodal Affective Computing}

\author{Ronghao~Lin\hspace{-1.0mm}\orcidRonghao{}\hspace{-1.0mm},
Ying~Zeng, 
Sijie~Mai\hspace{-1.0mm}\orcidSijie{}\hspace{-1.0mm}, 
and Haifeng~Hu\hspace{-1.0mm}\orcidHuhaifeng{} \hspace{-2.0mm}\IEEEmembership{,~Member,~IEEE}
\thanks{ This work was supported by the National Natural Science Foundation of China (62076262, 61673402, 61273270, 60802069). \protect\\
The authors are with the School of Electronics and Information Technology, Sun Yat-sen University, Guangzhou 510006, China. (E-mail: \{linrh7,zengy268,maisj\}@mail2.sysu.edu.cn, huhaif@mail.sysu.edu.cn).
 }
}

\markboth{Journal of \LaTeX\ Class Files,~Vol.~14, No.~8, August~2021}%
{Shell \MakeLowercase{\textit{et al.}}: A Sample Article Using IEEEtran.cls for IEEE Journals}

\IEEEpubid{0000--0000/00\$00.00~\copyright~2021 IEEE}

\maketitle

\begin{abstract}
In the pathway toward Artificial General Intelligence (AGI), understanding human's affection is essential to enhance machine’s cognition abilities. For achieving more sensual human-AI interaction, Multimodal Affective Computing (MAC) in human-spoken videos has attracted increasing attention. However, previous methods are mainly devoted to designing multimodal fusion algorithms, suffering from two issues: \textit{semantic imbalance} caused by diverse pre-processing operations and \textit{semantic mismatch} raised by inconsistent affection content contained in different modalities comparing with the multimodal ground truth. Besides, the usage of manual features extractors make they fail in building end-to-end pipeline for multiple MAC downstream tasks. To address above challenges, we propose a novel end-to-end framework named \textit{SemanticMAC} to compute multimodal semantic-centric affection for human-spoken videos. We firstly employ pre-trained Transformer model in multimodal data pre-processing and design Affective Perceiver module to capture unimodal affective information. Moreover, we present a semantic-centric approach to unify multimodal representation learning in three ways, including gated feature interaction, multi-task pseudo label generation, and intra-/inter-sample contrastive learning. Finally, SemanticMAC effectively learn specific- and shared-semantic representations in the guidance of semantic-centric labels. Extensive experimental results demonstrate that our approach surpass the state-of-the-art methods on 7 public datasets in four MAC downstream tasks.
\end{abstract}

\section{Introduction}

\IEEEPARstart{M}{ultimodal} Affective Computing (MAC) aims at predicting the sentiment polarity, emotion class, or behavioral intention by comprehensively integrating information from different modalities of speakers such as textual (utterance), acoustic (human voice) and visual (facial expression, head movement, body gesture) modality in a human-centric video  \cite{poria2020beneath,baltruvsaitis2018multimodal}. With the surge of human-spoken content on social media platforms, research on multimodal affective computing has become crucial in the community of multimodal learning \cite{poria2017review}. Considering various application purposes, multimodal affective computing is divided into diverse specific tasks, including multimodal sentiment analysis \cite{zadeh2018mosei,gkoumas2021makes,mao2022m}, multimodal emotion recognition \cite{busso2008iemocap, poria2019meld, geetha2024multimodal}, multimodal humor and sarcasm detection \cite{hasan2019urfunny,castro2019mustard}. 

\begin{figure}[htbp]
	\centering
		\includegraphics[scale=0.45]{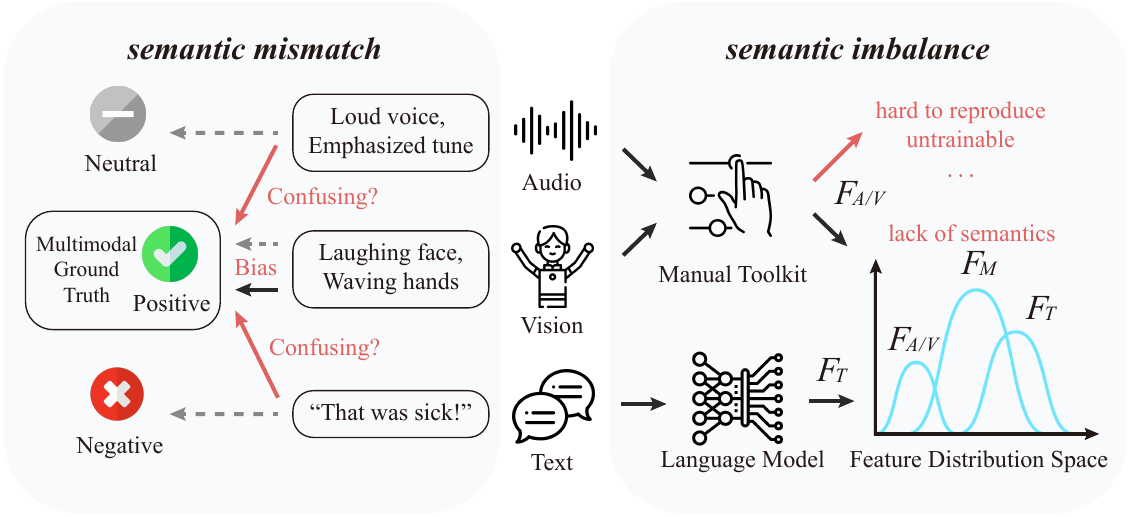}
        \vspace{-0.2cm}
	  \caption{The two main challenges in conducting multimodal affective computing from the perspective of semantic.}\label{figure0}
\vspace{-0.5cm}
\end{figure}

Affective computing are originated from conventional Natural Language Processing (NLP) tasks referring to understanding the affection contained in human-spoken utterances and conversations \cite{soleymani2017survey,poria2020beneath,lian2023gcnet}. The performance of affection-related algorithms highly relies on semantic information \cite{liang2024foundations} and are mostly improved by exploring the abundant semantic context embedded in language models. Nevertheless, immoderate reliance on language may easily overfit on subjective affective components, resulting in biased prediction \cite{zadeh2017tensor,wang2019words}. Thus, auxiliary features from other modalities, such as audio and image, are introduced to enhance affective understanding with multimodal learning \cite{poria2017review}. In previous MAC methods, unlike textual features learned by language models, acoustic and visual features are mostly extracted by manual pre-processing toolkit such as CMU-MultimodalSDK\footnote{\url{http://immortal.multicomp.cs.cmu.edu/}} \cite{tsai2019multimodal,zadeh2018multi,dai2020modality,gkoumas2021makes,hasan2021humor}, due to the information sparsity and inherent noise in audio and image. However, conducting multimodal learning with manual features may raise issues as shown in Figure \ref{figure0}. 

\IEEEpubidadjcol

On the one hand, the vague description of pre-processing causes the extraction of manual features hard to reproduce \cite{mao2022m}, introducing inevitable gap between training and inference stages for multimodal learning. Besides, the manual feature extractors such as COVAREP \cite{degottex2014covarep} and Facet \cite{iMotions} are untrainable, which brings difficulty in developing end-to-end multimodal learning pipeline and affects the generalization of the pre-trained models in various downstream scenarios.

On the other hand, due to the demand of semantic context for MAC task, the manual features such as facial landmarks for visual features and Mel-frequency cepstral coefficients for acoustic features, are not efficiently suitable for affection-related tasks. Lack of semantic information, such low-level features lead to poor embedding performance comparing with textual modality \cite{hazarika2020misa,mao2022m,lin2023mtmd} and bring semantic imbalance issue in multimodal learning. Since the scale of language models is increasing rapidly, the number of trainable parameters for other modalities are much smaller than the ones for textual modality for MAC models, which further exacerbates the semantic imbalance for various modalities.

\begin{figure}[htbp]
    \begin{center}
    \subfigure[MulT \cite{tsai2019multimodal} with Glove] {
     \label{fig_PRcurve_sota_glove}
     \centering
    \includegraphics[width=0.46\linewidth]{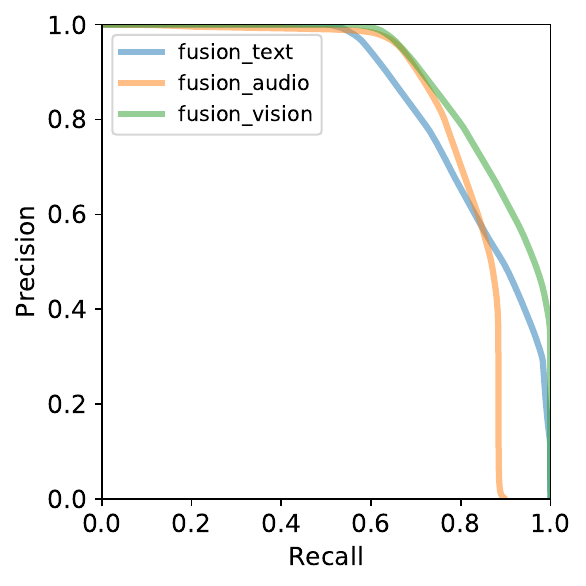}  
    }
    \subfigure[MulT \cite{tsai2019multimodal} with BERT] { 
     \label{fig_PRcurve_sota_bert1}
     \centering
    \includegraphics[width=0.46\linewidth]{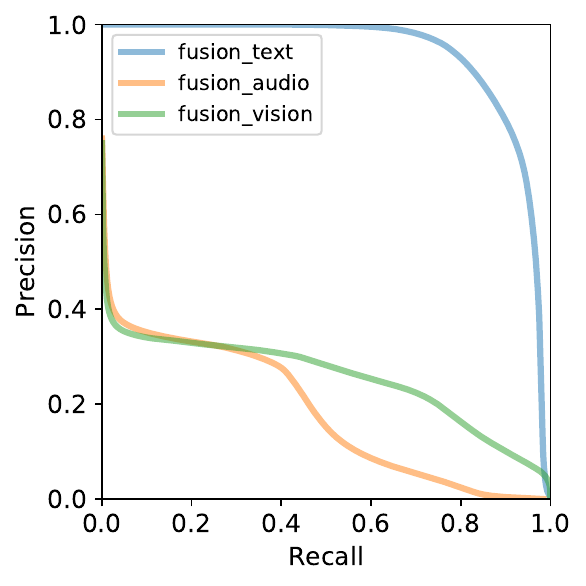}     
    }
    \\
    \subfigure[MISA \cite{hazarika2020misa} with BERT] { 
     \centering
    \includegraphics[width=0.46\linewidth]{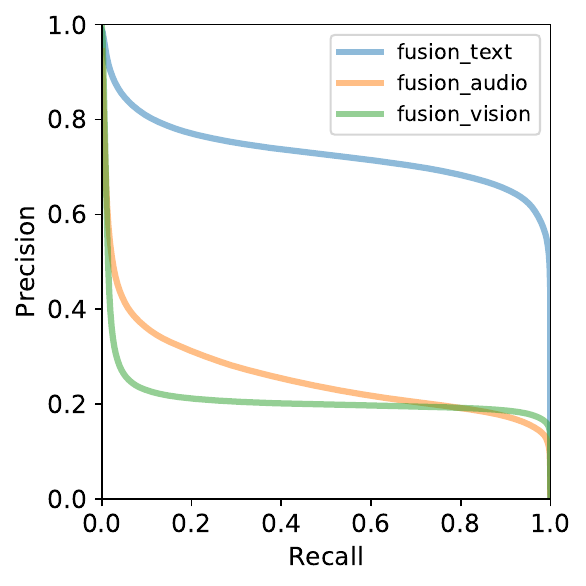}     
    }
    \subfigure[Self-MM \cite{yu2021learning} with BERT] { 
     \centering
    \includegraphics[width=0.46\linewidth]{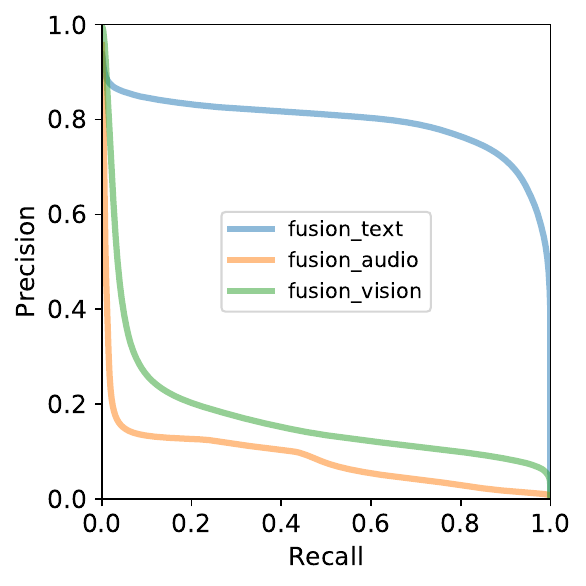}     
    }
    \\
    \subfigure[MMIM \cite{han2021improving} with BERT] { 
     \centering
    \includegraphics[width=0.46\linewidth]{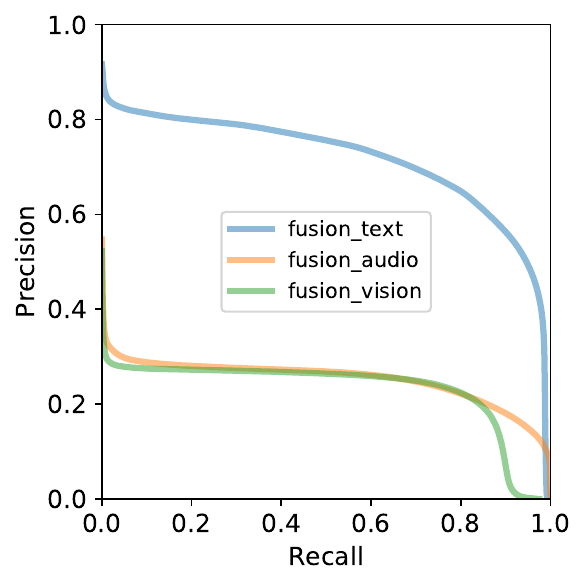}     
    }
    \subfigure[MissModal \cite{lin2023missmodal} with BERT] {  
     \label{fig_PRcurve_sota_bert2}
     \centering
    \includegraphics[width=0.46\linewidth]{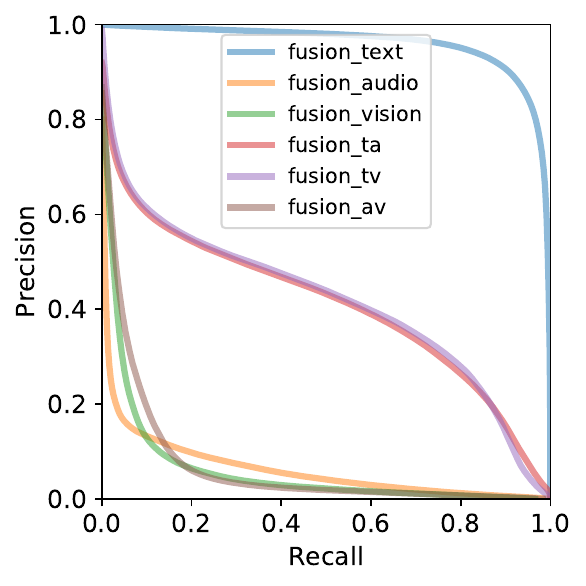}   
    }
    \caption{The PR curve of the fusion multimodal representations and the unimodal representations, including text, audio and vision modalities by state-of-the-art models training with Glove \cite{pennington2014glove} and BERT \cite{devlin2019bert} features on CMU-MOSEI dataset. Note that such PR curve is initially proposed as an evaluation metric for genrative models by Sajjadi \textit{et al.} \cite{sajjadi2018assessing} to formulate the relative probability densities of the distributions of real and generated data. }
    \vspace{-0.5cm}
    \label{fig_PRcurve_sota}
    \end{center}
\end{figure}

To better understand the issue of \textit{semantic imbalance} for different modalities, we visualize the contribution of the unimodal features for the fusion multimodal representations in Figure \ref{fig_PRcurve_sota}. Inspired by but diverse from \cite{sajjadi2018assessing,kynkaanniemi2019improved}, we compute the Precision-Recall (PR) curve for the feature distributions between the unimodal and multimodal representations, taking the representations from state-of-the-art multimodal sentiment analysis models \cite{tsai2019multimodal,hazarika2020misa,yu2021learning,han2021improving,lin2023missmodal} as examples. 

As shown in Figure \ref{fig_PRcurve_sota_glove}, we can observe that the manual acoustic and visual features contribute similarly when utilizing low-level textual features such as Glove \cite{pennington2014glove} which computes word vector based on global word co-occurrence counts statistics. However, when we substitute the textual features with BERT \cite{devlin2019bert} which embeds high-level semantic context by pre-trained language model in Figure \ref{fig_PRcurve_sota_bert1}-\ref{fig_PRcurve_sota_bert2}, the contributions of manual acoustic and visual features drop significantly to the multimodal representations compared with the one of textual features, no matter in which models. Although the existing fusion strategies \cite{gkoumas2021makes} may adjust the contributions of different unimodal representations adaptively, they fail in balancing the contributions of different modalities, mostly due to the inherent discrepancy of semantic abundance from various unimodal representations. 

Moreover, we remove features from each modality input in traversal manner as MissModal \cite{lin2023missmodal} to construct unimodal, bimodal and trimodal representations, and then compute the PR-curve among the distributions of these representations as shown in Figure \ref{fig_PRcurve_sota_bert2}. The bimodal representations with textual features contribute more than the representations with acoustic and visual features solely or both, which further indicates that the introduction of textual features can effectively increase the semantic information to the fusion multimodal representations. 

From the visualization in Figure \ref{fig_PRcurve_sota}, we can conclude that existing low-level manual acoustic and visual features are no longer appropriate for high-level textual features embedded by context-based language model. The difference of semantic abundance from various modalities causes the issue of semantic imbalance and affects the multimodal fusion process, leading to an urgent need of new solutions for unimodal feature extraction of acoustic and visual modalities. 

In addition, different modalities may bring diverse affective intensities or classes for MAC task \cite{zadeh2017tensor,dai2020modality,lin2023dynamically}, meaning that the affection semantics of various modalities may not remain consistent in the same video. Previous methods categorize unimodal features into modality-specific and -shared features to deal with such semantic inconsistency circumstance \cite{akhtar2019multi,hazarika2020misa,wu2021text,lin2023mtmd}. However, they utilize final multimodal ground truth labels to jointly supervise representations learning, confusing the training of modality-specific features with different affection as shown in Figure \ref{figure0}. We summarize this issue as \textit{semantic mismatch} raised by inconsistent semantics among unimodal features and the corresponding multimodal ground truth labels. Moreover, as interpreted in Du \textit{et al.} \cite{du2023uni}, multimodal joint training easily suffer from modality laziness, which makes the model neglect the learning of modality-specific features regardless of the paired features. Therefore, relying solely on annotated multimodal labels as the supervision is insufficient for multimodal learning \cite{yu2021learning}. Particularly for MAC task, it is crucial to explore the unimodal semantics contained in various modalities and enhance nuanced comprehension of fine-grained affection in multimodal learning, ensuring more precise prediction without bias \cite{yu2020chsims}. The case study in Table \ref{case_study_mismatch} further reveals that the MAC task require individual supervision signals to capture the affective semantic information for various modalities.

\begin{table}[htbp]
\centering
  \scriptsize
  \setlength{\tabcolsep}{3pt}
  \caption{Case study of SemanticMAC to tackle semantic mismatch. } 
  \label{case_study_mismatch}
  \scalebox{0.9}{\begin{tabular}{cc|c|c|cc}
    \toprule[1.5pt]
    \# & Modality Description & \makecell{Pseudo \\Label $p_*$} & \makecell{Ground \\Truth $y_{gt}$} & \multicolumn{2}{c}{\makecell{SemanticMAC \\Prediction $\hat{y}_*$}}\\
    \midrule[1.5pt] 
    \multirow{3}{*}[-17pt]{1} & \makecell[c]{``It's berry berry berry red and \\it's way too not good for me"} & \colorbox{red!45}{-1.006} & \multirow{3}{*}[-17pt]{\colorbox{black!20}{0.000}} & \colorbox{red!20}{-0.395} & \multirow{3}{*}[-17pt]{\colorbox{black!10}{-0.110}}\\
    & \makecell[c]{\includegraphics[width=1.5cm]{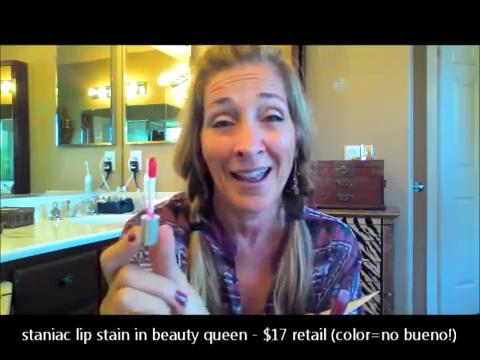} ... \includegraphics[width=1.5cm]{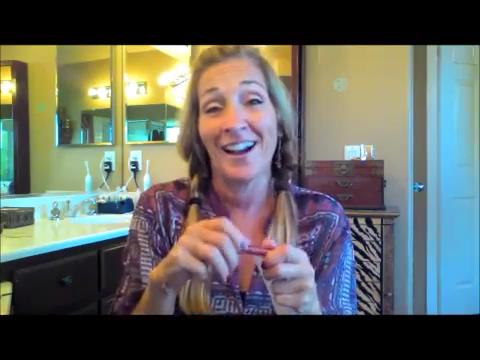}} & \colorbox{green!30}{1.016} & & \colorbox{green!30}{0.875} & \\
    & Soaring and emphatic tone & \colorbox{green!30}{0.902} & & \colorbox{green!40}{1.397} & \\ 
    \midrule[0.5pt]
    \multirow{3}{*}[-17pt]{2} & \makecell[c]{``Hannibal Lecter is one twisted character \\and this movie's all about him"} & \colorbox{red!20}{-0.330} & \multirow{3}{*}[-17pt]{\colorbox{green!45}{1.333}} & \colorbox{red!20}{-0.373} & \multirow{3}{*}[-17pt]{\colorbox{green!45}{0.936}}\\
    & \makecell[c]{\includegraphics[width=1.5cm]{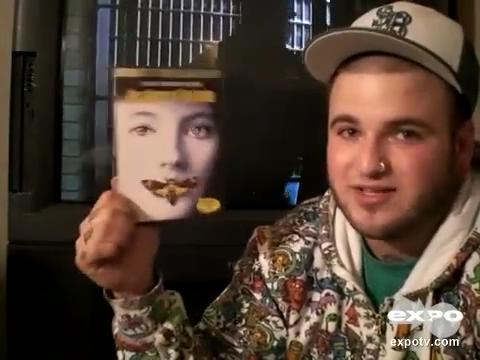} ... \includegraphics[width=1.5cm]{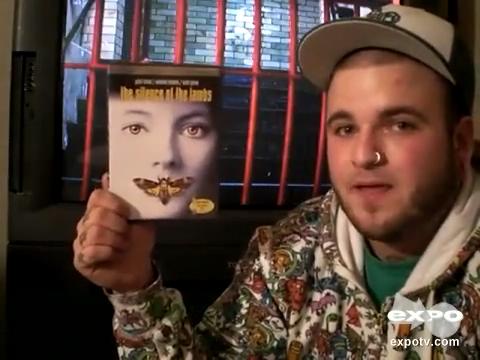}} & \colorbox{green!20}{0.870} & & \colorbox{green!30}{0.977} & \\
    & Excited and fast tone & \colorbox{red!10}{-0.143} & & \colorbox{red!30}{-0.988} & \\ 
    \midrule[0.5pt]
    \multirow{3}{*}[-13pt]{3} & \makecell[c]{``Rent this one"} & \colorbox{green!5}{0.146} & \multirow{3}{*}[-13pt]{\colorbox{red!20}{-0.667}} & \colorbox{black!20}{0.003} & \multirow{3}{*}[-13pt]{\colorbox{red!20}{-0.414}}\\
    & \makecell[c]{\includegraphics[width=1.5cm]{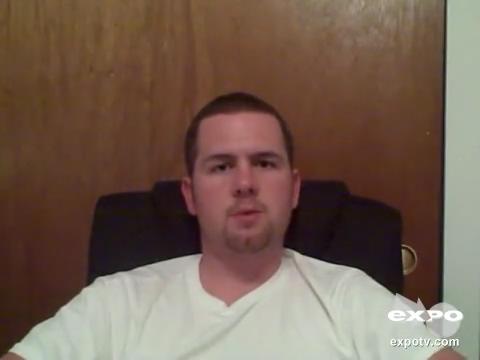} ... \includegraphics[width=1.5cm]{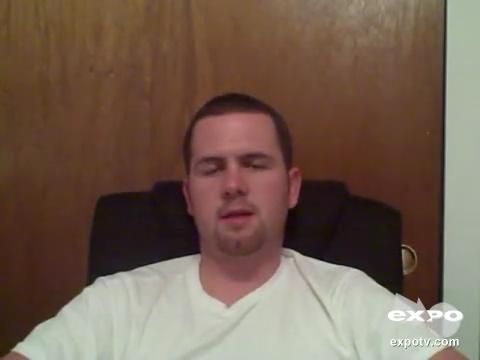}} & \colorbox{red!40}{-1.274} & & \colorbox{red!30}{-1.073} & \\
    & Depressed and lowering tone & \colorbox{red!30}{-1.080} & & \colorbox{red!25}{-0.802} & \\ 

    \midrule[0.5pt]
    \multirow{3}{*}[-17pt]{4} & \makecell[c]{``You really don't want to even \\mess with this movie"} & \colorbox{red!20}{\makecell[c]{Anger\\Disgust}} & \multirow{3}{*}[-17pt]{\colorbox{red!20}{Disgust}} & \colorbox{red!20}{Anger} & \multirow{3}{*}[-17pt]{\colorbox{red!20}{Disgust}}\\
    & \makecell[c]{\includegraphics[width=1.5cm]{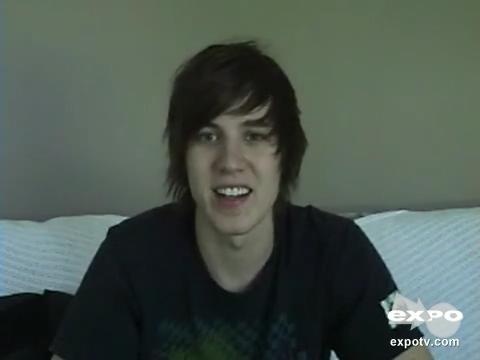} ... \includegraphics[width=1.5cm]{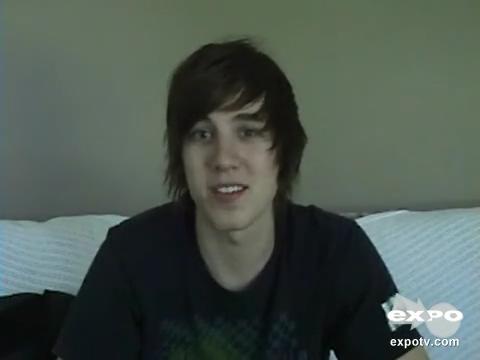}} & \colorbox{green!20}{Happy} & & \colorbox{green!20}{Happy} & \\
    & Negative and emphatic tone & \colorbox{red!20}{\makecell[c]{Disgust}} & & \colorbox{red!20}{Disgust} & \\ 
    \midrule[0.5pt]
    \multirow{3}{*}[-17pt]{5} & \makecell[c]{``You can choose to work with a \\transaction broker or a buyer's agent"} & \colorbox{red!20}{Sad} & \multirow{3}{*}[-17pt]{\colorbox{red!20}{\makecell[c]{Sad\\Fear}}} & \colorbox{black!20}{None} & \multirow{3}{*}[-17pt]{\colorbox{red!20}{\makecell[c]{Sad\\Fear}}}\\
    & \makecell[c]{\includegraphics[width=1.5cm]{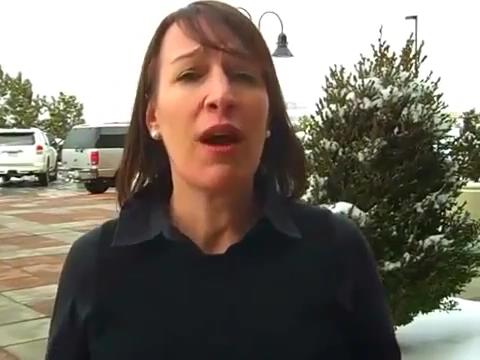} ... \includegraphics[width=1.5cm]{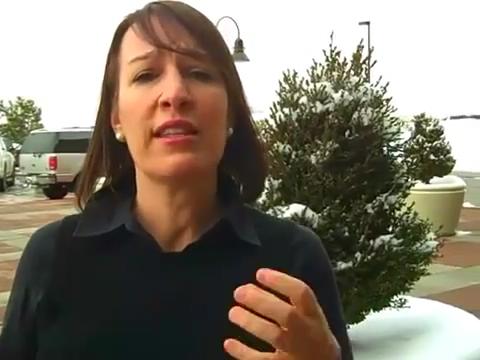}}& \colorbox{red!20}{\makecell[c]{Sad\\Fear}} & & \colorbox{red!20}{\makecell[c]{Sad\\Anger}} & \\
    & Uncertained and rhetorical tone & \colorbox{red!20}{Fear} & & \colorbox{red!20}{Fear} & \\ 
    \midrule[0.5pt]
    \multirow{3}{*}[-17pt]{6} & \makecell[c]{``But I really didn't like the apocalyptic \\ending, its just left me disappointed"} & \colorbox{red!20}{\makecell[c]{Sad\\Anger}} & \multirow{5}{*}[-8pt]{ \makecell[c]{\colorbox{green!20}{Surprise}\\ \colorbox{red!20}{\makecell[c]{Sad\\Anger\\Disgust}} }} & \colorbox{red!20}{\makecell[c]{Sad}} & \multirow{5}{*}[-8pt]{ \makecell[c]{\colorbox{green!20}{Surprise}\\ \colorbox{red!20}{\makecell[c]{Sad\\Disgust}} }}\\
    & \makecell[c]{\includegraphics[width=1.5cm]{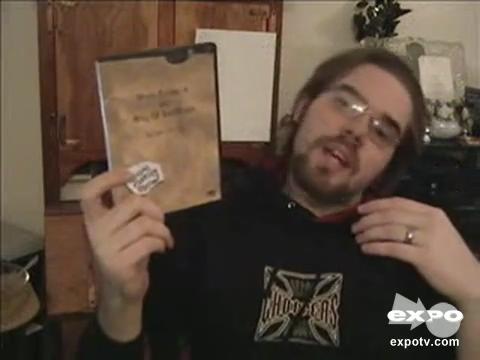} ... \includegraphics[width=1.5cm]{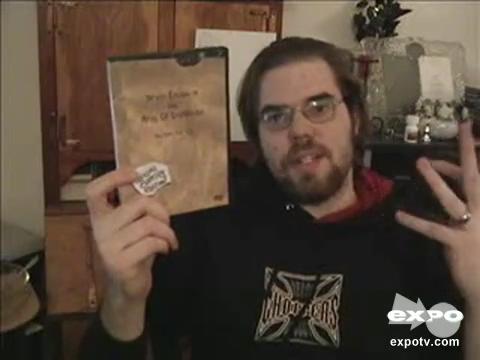}} & \colorbox{red!20}{\makecell[c]{Sad\\Disgust}} & & \colorbox{red!20}{\makecell[c]{Sad\\Disgust}}  & \\
    & Transition and definite tone & \colorbox{green!20}{Surprise} & & \colorbox{green!20}{Surprise} & \\ 
    \midrule[1.5pt]
  \end{tabular}}
\vspace{-0.5cm}
\end{table}

Aiming at addressing the challenges of \textit{semantic imbalance and mismatch}, we propose a novel \textbf{Semantic-centric Multimodal Affective Computing} framework, named \textbf{SemanticMAC}, to learn multimodal representations in the semantic space for various video-based MAC tasks in an end-to-end manner. Firstly, we utilize powerful pre-trained Transformer models \cite{vaswani2017attention} instead of manual features to extract unimodal features of different modalities from the raw videos. Such pre-processing operation ensures the end-to-end training and inference of the multimodal learning model. In order to reduce the modality heterogeneity and generalize to various scenarios for MAC task, we unify the embedding form for diverse input modalities according to the temporal sequences. Inspired by the thought of positional embedding \cite{kim2021vilt}, we utilize learnable frame embedding to denote videos with different frame lengths, which enhances the performance when dealing with varying length human-spoken videos. To further collect taffective information, we design a module named affective perceiver to process the features into fixed number of learnable tokens in the latent space, meanwhile filtering the noisy content contained in the generic acoustic and visual features. 

Then, we conduct Semantic-centric Gated Feature Interaction (SGFI) inside and among the unimodal features from various modalities by bridge attention and gated mechanism to extract specific- and shared-semantic representations. The former representations explore the intra-modal dynamics for affection-specific knowledge and the latter ones integrates the cross-modal commonality by reducing the modality gap, both of which enhance model's ability of affective perception and multimodal reasoning. Next, targeting at training representations with various affective content, we present Semantic-centric Label Generation (SCLG) to calculate specific- and shared-semantic labels for each sample from multimodal ground truth in a momentum-updated policy. The generated pseudo labels are served as weak supervision signals to guide the learning of specific- and shared-semantic representations in a multi-task training paradigm, alleviating the semantic mismatch issue of multimodal joint training. 

Besides, diverse with conducting contrastive learning among paired modalities \cite{radford2021learning,li2022clmlf,hu2022unimse}, we perform Semantic-centric Contrastive Learning (SCCL) for various modalities from the perspectives of intra- and inter-sample. The introduction of semantics in contrastive learning significantly improves the convergence of representations for both unimodal and multimodal sub-tasks. Specifically, the intra-sample one measure the similarity of features from different modalities in each sample, encouraging cross-modal interaction with the guidance of specific and shared semantics. While the inter-sample one is presented under the guidance of the sentiment intensity or emotion classes of multimodal representations, enabling affection-related cooperation in the multimodal fusion. Lastly, we calculate the multi-task losses supervised by the ground-truth and generated paseudo labels and the contrastive learning losses as the final optimization objective.

The main contributions of our paper can be summarized as:

\begin{itemize}
\item \textbf{A unified and novel end-to-end framework for MAC:} Focusing on the affective semantic of for textual, acoustic and visual modalities from the human-spoken video, we propose a novel framework named SemanticMAC to unify the learning process of multimodal representations and predict human's affective intensity in an end-to-end manner for multimodal affective computing (MAC) task. Rather than manual toolkit in previous methods, we utilize the pre-trained Transformer model and design the affective perceiver to extract unimodal features, which enable flexible pre-processing with various length video and address the issue of semantic imbalance.

\item \textbf{Semantic-centric representation learning approach:} According to the specific semantics inside each modality and the shared semantics across diverse modalities, we present Semantic-centric Gated Feature Interaction (SGFI) to capture intra- and cross-modal dynamics. Meanwhile, we introduce Semantic-centric Label Generation (SCLG) to generate weak supervision for specific- and shared-semantic representations respectively, which eases the semantic mismatch in the label space. We also conduct Semantic-centric Contrastive Learning (SCCL) to promote the interaction among modalities and across samples guided by semantics and affection information. 

\item \textbf{Achieving state-of-the-art performance:} Extensive experiments demonstrates the effectiveness of our approach on 7 public datasets for 4 MAC downstream tasks universally, including multimodal sentiment analysis, multimodal emotion recognition, multimodal humor and sarcasm detection.

\end{itemize}
 
\section{Related Work}
\subsection{Multimodal Affective Computing}
As a sub-field of multimodal learning, the key question of Multimodal Affective Computing (MAC) is summarized as how to extract semantic-rich unimodal features and effectively fuse the affective related information from each modality to learn multimodal representations \cite{baltruvsaitis2018multimodal,manzoor2023multimodality,lin2024adapt}. Therefore, developing pipeline to conduct MAC contains two main aspects: unimodal feature extraction and multimodal fusion \cite{poria2020beneath,poria2017review}. 

Compared with the traditional low-level hand-craft features \cite{ mikolov2013efficient,baltruvsaitis2018openface2,degottex2014covarep}, unimodal features extracted by deep learning based models have achieved impressive performance for diverse modalities when applied in different fields. Particularly, unimodal pre-trained models consist of Transformer \cite{vaswani2017attention}, such as BERT \cite{devlin2019bert} and GPT \cite{radford2018improving} for text in natural language processing, ViT \cite{dosovitskiy2021image} for image in computer vision, and HuBERT \cite{hsu2021hubert} for audio in speech processing, are capable of learning efficient unimodal representations and generalizing to various downstream tasks in the pre-train and fine-tuning paradigm. Additionally, multimodal fusion focuses on jointly integrating information from diverse modalities to perform affective prediction \cite{baltruvsaitis2018multimodal}. Gkoumas \textit{et al.} \cite{gkoumas2021makes} and Geetha \textit{et al.} \cite{geetha2024multimodal} have provided comprehensive surveys on the current multimodal fusion techniques for MAC, which have attained remarkable results while still suffered from the huge modality gap and the issues of semantic imbalance and mismatch.

The MAC task consists of multiple affective prediction downstream tasks, including 1) multimodal sentiment analysis \cite{zadeh2018mosei,gkoumas2021makes,mao2022m} to compute a continuous score as the sentiment polarity of utterance in a regression method; 2) multimodal emotion recognition \cite{busso2008iemocap, poria2019meld, geetha2024multimodal} to classify the emotion class of the utterance in monologue or conversation; 3) multimodal humor and sarcasm detection \cite{hasan2019urfunny,castro2019mustard} to identify whether the utterance contains the humorous or satirical intent.
Previous methods address single task according to distinctive forms of input data and objective functions. Differently in this paper, we present one unifying framework to effectively adopt these downstream tasks, providing a unique insight for future research.

\subsection{Attention Mechanism}
Current multi-head attention mechanism is mostly based on Transformer \cite{vaswani2017attention}, named self-attention, which presents normalized scaled dot-product among the input query, key and value from the same input sequence. Multiple variants of attention mechanism have been proposed to adapt in distinct scenarios, such as linear attention \cite{katharopoulos2020transformers} to reduce the inference computation from quadratic complexity into linear one, cross-attention \cite{tan2019lxmert} to process different input, and mutli-query attention \cite{shazeer2019fast} to decrease the model parameters and the key/value cache and so on. 

Multimodal learning with attention mechanism has been exploited extensively in previous researches \cite{xu2023multimodal}. Whisper \cite{radford2023robust} trains a robust speech recognition model by cross-attention with a large-scale web text-audio data as a weakly supervised datasets in a multi-task training approach. Flamingo \cite{alayrac2022flamingo} presents perceiver resampler to convert varying-size large feature maps to fixed visual tokens and interact these tokens with textual data by masked attention layers in the vision-language pre-training. However, most of them leverage the attention mechanism to construct cross-modal connection instantly, ignoring the different interaction manners and various supervision of the fine-grained features in the semantic space. 

\subsection{Contrastive Learning}
Contrastive learning focuses on dividing the samples into positive and negative pairs sets and adjusting the similarity of the corresponding representations \cite{hadsell2006dimensionality}. The most popular form of contrastive loss function is InfoNCE, which is utilized to encode underlying shared latents by maximizing the lower bound of the mutual information \cite{van2018representation}. As a pretext task, contrastive learning is initially adopted at unimodal models in an unsupervised manner \cite{he2020momentum,chen2020simple,gao2021simcse}, and then extended to supervised methods and multimodal models due to its great effectiveness and generality. Supcon \cite{khosla2020supervised} leverages label information to conduct contrastive learning in a fully-supervised setting. Recent works such as CLIP \cite{radford2021learning}, ALIGN \cite{jia2021scaling} and wav2vec2.0 \cite{baevski2020wav2vec} and so on have claimed the better cross-modal alignment performance with contrastive objectives. Particularly, ImageBind \cite{girdhar2023imagebind} extend contrastive learning into the joint embedding space across six modalities. Nevertheless, most of them lack exploration into multimodal fusion and reveal significant modality gap \cite{liang2022mind}, which is imperatively needed to be addressed.

\begin{figure*}[htbp]
	\centering
		\includegraphics[scale=0.5]{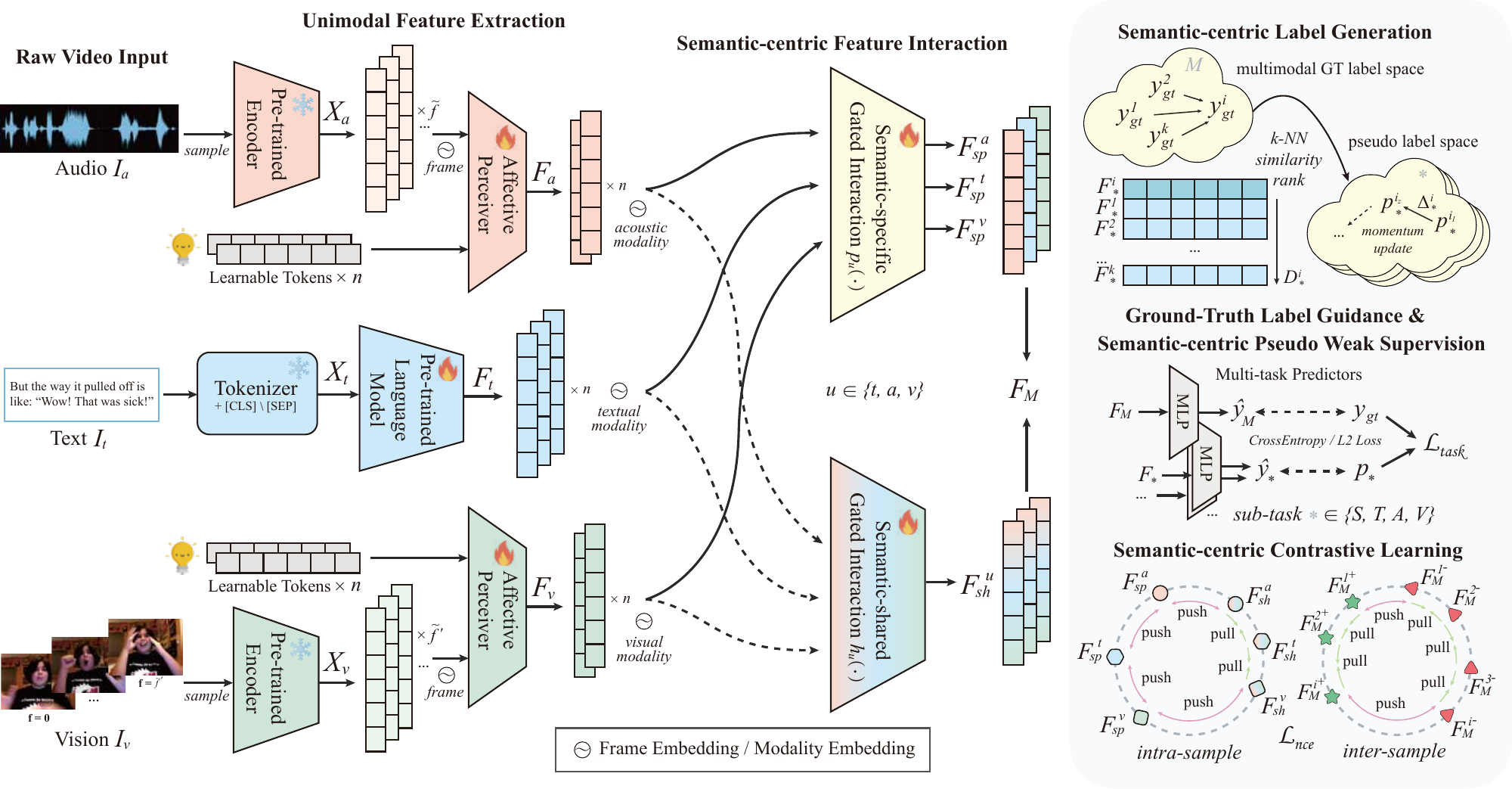}
	  \caption{The overall architecture of the proposed SemanticMAC. Note that the frame embeddings and modality embeddings are updated during the stage of training while then fixed and generalized into the downstream inference.}\label{figure1}
\end{figure*}

\section{Methodology}
The proposed SemanticMAC is presented in detail in this section. We first define the input and output of MAC task and clarify the corresponding notations. Then we introduce the end-to-end architecture of the proposed framework. Next, we put forward the extraction process of unimodal features for different modalities. Following the semantic-centric thought, the module of gated feature interaction and the strategy of label generation are described additionally. Finally, we formulate the total optimization objective with the semantic-centric contrastive learning loss and the individual task prediction losses.
\subsection{Problem Definition}
Multimodal Affective Computing (MAC) concentrates on learning efficient representations to conduct various regression or classification tasks for affective analysis from the multimodal signals contained in a human-spoken video. To unify diverse downstream MAC tasks, we formulate the multimodal input of the raw videos as $I_u\in\mathbb{R}^{\ell_u\times c_u}$, where $\ell_u$ denotes the temporal length of utterance sequence and $c_u$ denotes various contents of unimodal signal at the sampled timestep of the video. Particularly, since each video clip contains at least an utterance spoken by one human with facial expression, head movement and body gesture, $u\in\{t,a,v\}$ represents the textual, acoustic and visual modalities respectively \cite{poria2017review}. According to the semantics contained in various modalities, the proposed SemanticMAC processes each modality of the raw multimodal data to unimodal representations $F_u$ and then integrates the affective information into multimodal representations $F_M$ by cross-modal interaction in the semantic space. Lastly, the task predictor utilize the final mutlimodal representations to output $\hat{y}_M$, which serves as the sentiment scores in regression task or as the emotion classes in recognition and detection task.

\subsection{Architecture Overview}
The architecture of the proposed SemanticMAC processing raw video in an end-to-end manner is depicted as Figure \ref{figure1}. Aiming at avoiding the issue of semantic imbalance from root, we firstly take pre-trained Transformer models instead of manual toolkits to pre-process unimodal data into embeddings with consistent form for various modalities. The unimodal embeddings $X_u$ are multiple tokens in $\mathbb{R}^{\tilde{f_u}\times d_u}$ where $\tilde{f_u}$ denotes the numbers of tokens and $d_u$ denotes the representation dimension of modality $u\in\{t,a,v\}$. Then, for acoustic and visual modalities, we design the affective perceiver to integrate affective information contained in the generic unimodal embeddings and transfer the knowledge into multiple learnable tokens $F_a$ and $F_v$ with fixed length, which enable the flexible handling of videos with different lengths at the same time. Besides, the acoustic and visual features are summed with learnable frame embedding $E_{fr}$ to attend by relative temporal order of various frames of the video. While for textual modality, the pre-trained language model is utilized to learn the affective textual representation $F_t$. Note that we freeze the pre-processed encoders for acoustic and visual modalities and the tokenizer for textual modality during the training stage while update the parameters of affective perceivers and language model as fine-tuning paradigm. Next, to make the model distinguish different modalities in latter modules, the unimodal representations $F_u$ are attached with learnable modality embeddings $E_{md}$ according to the corresponding type of modality. In addition, we conduct semantic-centric feature interaction among various unimodal representations to learn semantic-specific representations $(F^t_{sp}, F^a_{sp}, F^v_{sp})$ and semantic-shared representations $(F^t_{sh}, F^a_{sh}, F^v_{sh})$ for each modality, which further address semantic imbalance issue induced by overfitting on the dominant modality. Specifically, we design the interaction mechanism as gated multi-head intra- and cross-attention to efficiently capture intra-modal dynamics and explore cross-modal commonality. To focus on the learning of query modality at one time, we measure the attention score with multi-query setting in each interaction. Additionally, to reduce the impact of the modality gap introduced by modality heterogeneity, we utilize a set of bridge tokens to interact the information between query and key modalities with massively diverse distributions. Lastly, we concatenate the semantic-specific representations $(F^t_{sp}, F^a_{sp}, F^v_{sp})$ and semantic-shared representations $F_{sh}$ and then project them into multimodal representations $F_M$, which are fed to the task predictor to output the final affective prediction $\hat{y}_M$.

Targeting at the issue of semantic mismatch raised by various contents of each modality, we tend to utilize different semantic-centric labels as the supervision for different features in a multi-task training manner, which competently guides the learning of unimodal and multimodal representations in the semantic space. According to semantic attributes, we divide the training of representations into five sub-tasks denoted as $*\in\{M,S,T,A,V\}$, including the sub-tasks of multimodal representations $(M)$, semantic-shared representations $(S)$ and semantic-specific representations for each modality $(T,A,V)$. Most datasets only manually annotate the multimodal ground-truth labels $y_{gt}$ for multimodal representations $F_M$ in sub-task $M$ \cite{yu2020chsims,gkoumas2021makes}. Due to this, we consider to generate pseudo labels $p_*$ according to the fine-grained level of semantics for other representations $(F^t_{sp}, F^a_{sp}, F^v_{sp}, F_{sh})$ as a weakly-supervised strategy compared with the ground-truth annotations. By calculating the similarity ranking matrix for each type of representation in the feature space, the pseudo labels are then generated by scaling and shifting the corresponding ground-truth labels of k-nearest neighborhood samples. Besides, we stabilize the generation process of the pseudo labels in a momentum-based updating policy as the training epoch increases. Supervised by the ground truth labels and pseudo labels, the multi-task predictors take the unimodal and multimodal representations as the input and output the affective prediction $\hat{y}_*$ for each sub-task. Moreover, we perform semantic-centric contrastive learning in the level of intra-sample and inter-sample at the unit hypersphere \cite{wang2020understanding} to further enhance the convergence of multimodal representations learning. The former one pulls closer the semantic-shared representations of all modalities inside the same video sample and pushes away the semantic-specific representations for each modality, which encourage the decoupling of semantic information for unimodal features. While the latter one constructs positive and negative pairs based on the ground-truth affective category for the multimodal representations among different samples. More technical details are introduced in the following subsections.

\subsection{Unimodal Feature Extraction}
To unify the pre-processing of various modalities, we adopt Transformer-based models to extract unimodal features. As shown in Figure \ref{figure1}, the text data $I_t$ are firstly processed to tokens $X_t$ by tokenizer according to the specific language models \cite{devlin2019bert,yang2019xlnet,reimers2019sbert} in particular downstream task. Note that our framework is suitable for various language model, which is latter validated in the experiments. Then, we utilize the pre-trained language model to learn the textual representations  $F_t$, which is formulated as:
\begin{equation}
\begin{aligned}
    X_t &= Tokenizer(I_t) \in\mathbb{R}^{\tilde{f_t}}\\
    F_t &= LanguageModel(X_t;\theta_t)\in\mathbb{R}^{\tilde{f_t}\times d_t}
\end{aligned}
\end{equation}
The pre-trained language model are set in a fine-tuning paradigm where the parameters $\theta_t$ are updated during the training stage.

While for the audio data $I_a$ and video data $I_v$, where the general upper limit of the micro-expression duration is observed as 1/2 seconds \cite{yan2014casme}, we uniformly sample the audio and vision stream at 2 frames per second, efficiently reducing the input data volume and the model inference time. Next, the sampled audio and video frames are directly fed into the frozen pre-trained encoders of ImageBind \cite{girdhar2023imagebind} to jointly learn acoustic embedding $X_a$ and visual embedding $X_v$, which stack all [CLS] tokens from each sampled frame of the stream, formulated as:
\begin{equation}
\begin{aligned}
I'_u &= UniformSample(I_u)\in\mathbb{R}^{\tilde{f_u}\times c_u},\ \ u\in\{a,v\} \\
X_u^f &= ImageBind(I'_u;\theta_u)[CLS]\in\mathbb{R}^{d_u}, \ \  f\in[1,\tilde{f_u}] \\
X_u &= Stack[X_u^1 \dots X_u^i]\in\mathbb{R}^{\tilde{f_u}\times d_u}
\end{aligned}
\end{equation}
where $\theta_u$ denotes the encoders parameters and the [CLS] token are taken as the global embedding to aggregate the contents of each frame \cite{girdhar2023imagebind,radford2021learning,dosovitskiy2021an}. Note that we leverage the power of ImageBind for its excellent performance in aligning different multimodal data in the joint embedding space \cite{su2023pandagpt}.

\begin{figure}[hbp]
	\centering
		\includegraphics[scale=0.6]{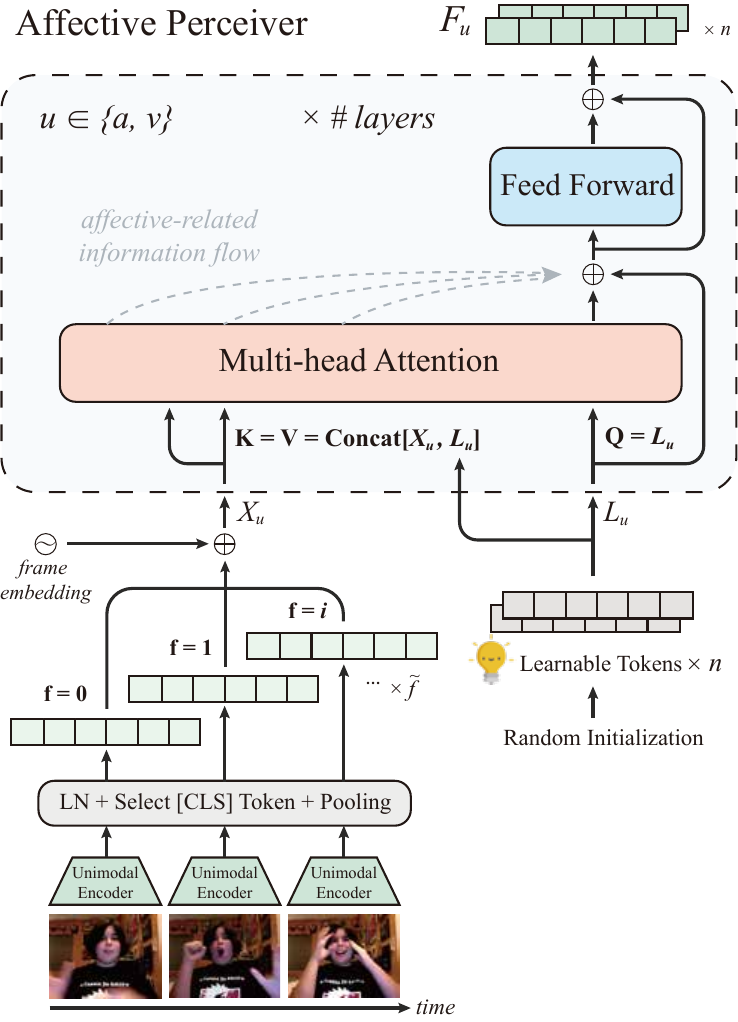}
	  \caption{The designed Affective Perceiver to learn affective unimodal features of acoustic and visual modalities. }\label{figure2}
\end{figure}

Although ImageBind has been proved effectively in multimodal alignment, the extracted unimodal embeddings are coarse-grained and generic in the embedding space, containing massive task-unrelated noise and affective-irrelevant information. Besides,  directly utilize [CLS] embeddings as the unimodal representations cause the model lack of temporal interaction for each modality. Thus, we design an extra module named Affective Perceiver to further learn fine-grained unimodal features and explore affective dynamics by interacting the [CLS] embeddings across frames as shown in Figure \ref{figure2}.

For acoustic and visual modalities, given video stream with $\tilde{f_u}$ frames, the unimodal embeddings extracted by the corresponding unimodal encoders of ImageBind are represented as $X^f_u$, where $u\in\{a,v\}$ and $f\in[1,\tilde{f_u}]$. Firstly, as the positional embedding in language model \cite{devlin2019bert}, we sum the unimodal embeddings $X_u$ with learnable frame embeddings $E_{fr}\in\mathbb{R}^{d}$ to increase relative temporal order to the module when conducting cross-frame attention, represented as: 
\begin{equation}
X_u = X_u + E_{fr}, \ \ u\in\{a,v\}
\end{equation}

Then, we innovate unimodal learnable tokens $L_u\in\mathbb{R}^{n\times d_u}$ with fixed length $n$ for individual modality aiming at collecting affective features in the generic unimodal embeddings. Due to the excellent performance of attention mechanism \cite{vaswani2017attention}, we design a multi-layer Transformer-based module named as Affective Perceiver by adopting multi-head attention (MHA) and feed forward network (FFN) in each layer. For $u\in\{a,v\}$, the Affective Perceiver gradually encourages the affective information of the unimodal embeddings $X_u$ to flow to the learnable tokens $L_u$. By constructing query, key and value as $Q=L_u$, $K=V= Concat[X_u,L_u]$, the computations of each Affective Perceiver layer are formulated as follows:   
\begin{equation}
\begin{aligned}
L_u &= MHA(LN(Q,K,V)) + L_u \\
L_u &= FFN(LN(L_u)) + L_u \\
\end{aligned}
\end{equation}
where layer normalization (LN) and residual connections are employed around each of the sub-layers. Note that $L_u$ is initialized randomly and the output of the last layer is taken as the unimodal representations $F_u\in\mathbb{R}^{n\times d_u}$. The effectiveness of such fixed number of leanrbale tokens in content abstraction for various modalities have been proved in recent researches \cite{jaegle2021perceiver,alayrac2022flamingo,li2023blip2}. Moreover, the unimodal learnable tokens $L_u$ in the Affective Perceiver can not only integrate the most useful information for downstream tasks while removing irrelevant noise, but empower the model with the ability to align acoustic and visual modalities with different language model in the feature space. With the introduction of Affective Perceiver, the proposed framework is capable of processing video with various frame length and extracting affective unimodal features competently, which further address the issue of semantic imbalance as shown in Figure \ref{figure0}.

\subsection{Semantic-centric Feature Interaction}
After the extraction of unimodal features, the essential question of multimodal learning has become how to interact various type of information from different modalities and conduct multimodal fusion with huge modality gap. In the perspective of semantic, we decouple the feature space into semantic-specific and semantic-shared features, where the former features focus on modal-specific semantic information according to the contents of diverse modalities while the latter ones integrates the invariant commonalities among all modalities. Such feature disentanglement strategy is intuitive and works successfully with theoretical interpretability \cite{hazarika2020misa,wu2021text,liang2023multiviz}. Diverse from previous researches, we propose Semantic-centric Gated Feature Interaction (SGFI) to learn semantic-specific and -shared representations by the designed bridge attention and gated mechanism, which effectively transfer intra- and cross-modal knowledge through bridge tokens and filter the irrelevant features by weighted activation layer.
\begin{figure}[htbp]
	\centering
		\includegraphics[scale=0.5]{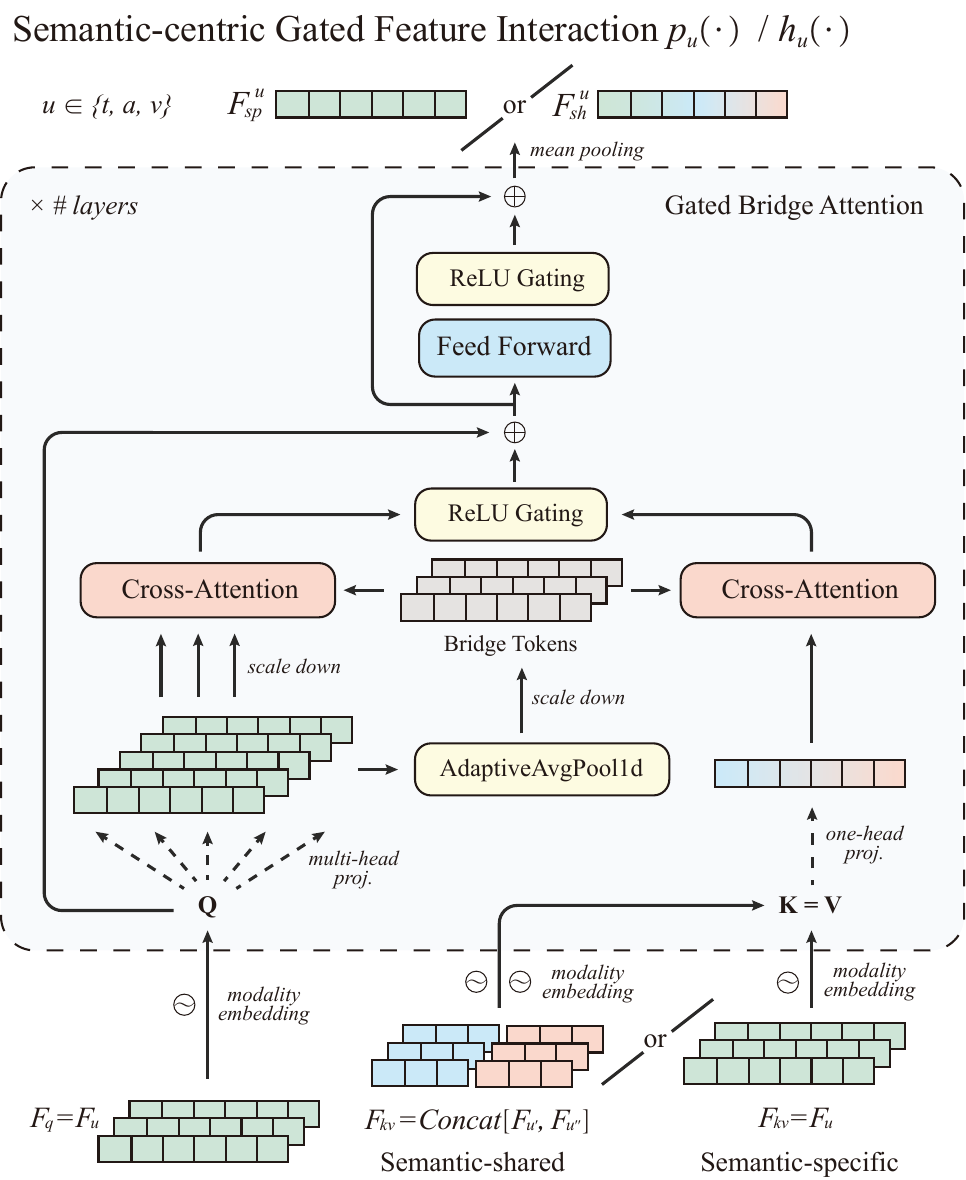}
	  \caption{The proposed Semantic-centric Gated Feature Interaction module.}\label{figure3}
\end{figure}

As shown in Figure \ref{figure3}, inspired by VilT \cite{kim2021vilt}, each unimodal representation $F_u$ is firstly summed with a learnable modality embedding $E_{md}\in\mathbb{R}^{d}$, which indicates the modality type for the module to distinguish corresponding representation in latter interaction, which is formulated as:
\begin{equation}
F_u = F_u + E_{md}, \ \ u\in\{t,a,v\}
\end{equation}

Similar as self-attention \cite{vaswani2017attention}, cross-attention has been proved competent in aligning different input data as query and key/value, respectively \cite{li2021align,rombach2022high,jaegle2022perceiver}. However, due to the huge modality gap and delicate modality relationship, the interaction in multimodal fusion for multimodal affective computing is far more complicated than simply multimodal alignment \cite{gkoumas2021makes}. Therefore, we improve the cross-attention mechanism in three ways for SGFI module, named Gated Bridge Attention (GBA), to adapt at the complex multimodal fusion:

\begin{itemize}
\item[1)] \textbf{Multi-query Attention:} We adopt multi-query attention \cite{shazeer2019fast,ainslie2023gqa} to primarily excavate various semantics inside query vectors, which accelerate the convergence of mutlimodal learning and lower the memory-bandwidth requirements concurrently. Specifically, we utilize multi-head projection $W^x_h$ for query vectors while maintain a single head of key/value vectors which share the same weights in the linear projection $W^y$ for each head of query vectors, formulated as:
\begin{equation}
\begin{aligned}
Q_h &= QW^x_h \in\mathbb{R}^{n\times d_{head}}, h\in[1, head] \\
K_h &= V_h = KW^y = VW^y \in\mathbb{R}^{n\times d_{head}}
\end{aligned}
\end{equation}
where $head$ denotes the number of heads, $d_{head}=d_c/head$ denotes the dimension of each head and $d_c$ is set as the common dimension for each representation.

\item[2)] \textbf{Bridge Token:} Aiming at bridging the modality gap among various modalities in the semantic space and conducting efficient feature interaction, we introduce Bridge Tokens with fixed $m$ tokens $(m<n)$ as bottleneck to restrict the intra- and cross-attention flow, inspired by the thought of information bottleneck \cite{alemi2017deep,nagrani2021attention,han2023agent}. The Bridge Tokens $B$ are obtained by aggregating features in adaptive average pooling based on semantics from query vectors:
\begin{equation}
\begin{aligned}
    Q'&=Concat[Q_1 \dots Q_{head}] \in\mathbb{R}^{n\times d_c} \\
    K'&= V' = Repeat(K_h) \in\mathbb{R}^{n\times d_c} \\
    B &= AdaptiveAvgPool(Q') \in\mathbb{R}^{m\times d_c}
\end{aligned}
\end{equation}

Then, scaling down by $\sqrt{d_c}$, the attention matrix is computed as:
\begin{equation}
BridgeAttn(Q,K,V)=Softmax(\frac{(Q'B^T)(BK')}{\sqrt{d_c}}V')
\end{equation}

\item[3)] \textbf{Gated ReLU:} To filter the redundancy according to the semantic of individual representations, we adopt the gated mechanism between each attention and feed forward sub-layer by Rectified Linear Unit (ReLU) \cite{fukushima1969visual}, which has been proved suitable for Transformer models due to its activation sparsity and inference efficiency \cite{li2023lazy,mirzadeh2024relu}. Thus, for $u\in\{t,a,v\}$, the computation in GBA is finally formulated as:
\begin{equation}
\begin{aligned}
F^u &= ReLU(BridgeAttn(Q,K,V)) + F^u \\
F^u &= ReLU(FFN(F^u)) + F^u \\
\end{aligned}
\end{equation}

\end{itemize}

The SGFI module is conducted by stacking multiple GBA attention layers and outputs semantic-centric representations according to the input modality, which are denoted as $p_u(\cdot)$ for semantic-specific feature interaction and $h_u(\cdot)$ for semantic-shared feature interaction. 

On the one hand, to capture intra-modal dynamics and filter affective-unrelated noise, we take unimodal representations from the same modality to construct the input query and key/value for the SGFI module, which are denoted as $Q=K=V=F_u$. Then, the semantic-specific representations $F^u_{sp}$ can be computed as:
\begin{equation}
    F^u_{sp} = AvgPool(p_u(F_u))\in \mathbb{R}^{d_c}, \ \ u\in\{t,a,v\}
\end{equation}

On the other hand, to effectively fuse knowledge among different modalities and incorporate the affective commonalities, given the input query as $Q=F_u$ from arbitrary modality, we set the key/value as $K=V=Concat[F_{u'}, F_{u''}]$ which is the concatenation of the other unimodal representations. Thus, the semantic-shared representations $F_{sh}$ are formulated as:
\begin{equation}
\begin{aligned}
    F^u_{sh} &= AvgPool(h_u(F_t, F_a, F_v)) \in \mathbb{R}^{d_c} \\
    F_{sh} &= Concat[F^t_{sh},F^a_{sh},F^v_{sh}] \in \mathbb{R}^{3d_c}
\end{aligned}
\end{equation}

Finally, to summarize the semantic-specific and -shared information from various modalities, the multimodal representations $F_M$ are formulated as:
\begin{equation}
    F_M = Concat[F^t_{sp}, F^a_{sp}, F^v_{sp}, F^t_{sh}, F^a_{sh}, F^v_{sh}] \in \mathbb{R}^{6d_c}
\end{equation}

\subsection{Semantic-centric Label Generation}
For the learning of various semantic-specific representations $F^u_{sp}$ and semantic-shared representations $F_{sh}$, the supervision should be produced according to the semantics information. However, due to the absence of unimodal labels in most datasets, most of previous work \cite{hazarika2020misa,wu2021text} directly utilize the multimodal ground truth labels to supervise the learning process of features with various semantic, which are essentially contrary with the thought of disentangled representation learning. Besides, the affection expressed through single modality can be quite diverse, which is concluded as the semantic mismatch issue as shown in Figure \ref{figure0}. Aiming at addressing this issue, we present Semantic-centric Label Generation (SCLG) to construct pseudo label space based on semantics as the weak supervision strategy to improve the learning of semantic-centric representations.

Specifically, we deem the learning processes of representations $F_*$ with various semantics as distinct sub-tasks $*\in\{S,T,A,V\}$, which denote the sub-task of semantic-shared representations $F_{sh}$ and semantic-specific representations $F^u_{sp}$ for textual, audio and visual modalities. Each subtask should be trained under the guidance of the corresponding semnatic-centric pseudo labels. Inspired by Yu \textit{et.al}\cite{yu2021learning}, the semantic-centric labels $p_*$ are assumed to share the distribution space with multimodal ground truth labels $y_{gt}$. Thus, we utilize the common semantics contained in the representations across various samples and their ground truth labels to generate the pseudo specific- and shared-semantic labels as shown in Figure \ref{figure1}.

Given a query of representations $\mathcal{B}=\{F^i_*\}^B_{i=1}$, we conduct k-Nearest Neighbor (k-NN) algorithm to find the $K$ most nearest samples $\{F^k_*\}^K_{k=1}(K<B)$ for each representation $F^i_*$ by comparing the similarity in the feature space and then output the euclidean distance matrix $D_*$ between each sample and the nearest samples, denoted as:
\begin{equation}
\begin{aligned}
    \{F^k_*\} &= k\text{-NN}(F^i_*; F^1_* \dots F^B_*) \in \mathbb{R}^{d_c}, \ i\in[1,B], k\in[1,K] \\
    D_* &= (D_*^{ik}), \ \text{where } D_*^{ik} = \sqrt{\frac{1}{d_c}\sum^{d_c}_{j=1}(F^{i}_{*j}-F^{k}_{*j})^2}
\end{aligned}
\end{equation}
where the dimension of the representations $d_c$ is utilized as a scaling factor to mitigate the adverse effect of excessive distance. The distance matrix $D_*$ represents the similarity of various representations, indicating the relationship among different samples at the level of specific or shared semantics.

For each sub-task, to transfer the knowledge of multimodal ground truth labels $y_{gt}$ to the semantic-centric pseudo labels $p_*$, we design a scaling map to control the transferring magnitude related to semantics abundance and a shifting map to decide the direction and value of the label movement $\Delta$. Therefore, we intuitively construct pseudo labels by considering the distance matrix $D_*$ in the Gaussian potential kernel form function \cite{cohn2007universally} as the scaling map, and the difference between multimodal labels $y^k_{gt}$ and $p_*$ of the corresponding nearest samples as the shifting map, which is formulated as: 
\begin{equation}
    \Delta^i_* = \frac{1}{K} \sum^K_{k=1} \underbrace{\exp_{}^{-\omega\cdot D^{ik}_*}}_{scale} \cdot \ (\underbrace{y^k_{gt}-p^i_*}_{shift})
\end{equation}
where $\Delta^i_*$ denotes the varying value for pseudo label $p^i_*$ of sample $i$ in sub-task $*$. Moreover, the pseudo labels are initialized as the corresponding multimodal ground truth labels and updated in a momentum manner by combining the computed movement and history values with the increasing of training epochs $z$, represented as:
\begin{equation}
\begin{aligned}
    p^i_*|^0 &= p^i_*|^1 = ... = p^i_*|^r = y^i_{gt},  \ \  r\geq 1\\
    p^i_*|^{z} &= \frac{z-1}{z}p^i_*|^{z-1} + \frac{1}{z} p^i_*|^{z}\\
    &=\frac{z-1}{z}p^i_*|^{z-1} + \frac{1}{z} (p^i_*|^{z-1}+\Delta^i_*|^{z})\\
    &= p^i_*|^{z-1} + \frac{1}{z} \Delta^i_*|^{z}, \ \ z>r
\end{aligned}
\end{equation}
where the momentum-based updating policy is intended to relieve the fluctuations caused by noisy samples and the updating process of pseudo labels is started after the $r$th epoch for more stable label generation and better convergence. 

For each subtask, the specific-semantic labels are generated to reveal the intra-modal connections among different samples while the shared-semantic labels are expected to show the inter-modal commonality. Comparing with the multimodal ground truth labels, the generated pseudo labels are used to guide the learning of various semantic-centric representations in a weakly-supervised manner. Note that the semantic-centric pseudo labels are allowed to be zero for individual samples when there are few unimodal features or rare paired features related to the downstream prediction \cite{du2023uni,liang2024foundations}. 

\subsection{Semantic-centric Contrastive Learning}
To promote the disentanglement of semantics and encourage the feature interaction of unimodal and multimodal sub-tasks, we conduct Semantic-centric Contrastive Learning (SCCL) among various modalities inside and across different samples. Previous works \cite{radford2021learning,xu2021videoclip} utilizes cross-modal contrastive learning directly on unimodal representations suffering from the huge modality gap \cite{liang2022mind}. Diversely, SCCL is presented in the perspective of intra- and inter-sample at the semantic space, where the modality gap has been mitigated by SGFI.

As suggested by Wang \textit{et al.} \cite{wang2020understanding}, we firstly employ L2-normalization on all representations for both intra- and inter-sample contrastive learning to restrict the contrastive learning space on the unit hypersphere, as shown in Figure \ref{figure1}. We implement the intra-sample contrastive learning among semantic-specific representations $F^u_{sp}$ and semantic-shared representations $F_{sh}$ for $u\in\{t,a,v\}$, which efficiently decouple the semantic-specific features and the consistent information contained in each modality $u$. Given $\{F^u_{sp}, F^u_{sh}\}$ from each sample $i$, the goals are pushing away $F^u_{sp}$ and pulling closer $F^u_{sh}$ from different modalities, and pushing apart $F^u_{sp}$ and $F^u_{sh}$ according to the semantics inside the corresponding representations. Thus, we construct positive pairs as $\{F^u_{sh};F^{u'}_{sh}\}$, and the negative pairs as $\{F^u_{sp};F^{u'}_{sp}\}$ and $\{F^u_{sp};F^{u'}_{sh}\}$, and adopt dot-product similarity between the query and key in the pairs, formulated as:
\begin{equation}
\begin{aligned}
    &sim(F_{query}, F_{key})  = \frac{1}{2} (F_{query} \cdot F_{key}^T + F_{query}^T \cdot F_{key})  \\
    &S^+ = \sum_{u\neq u'}\exp({sim(F^u_{sh}, F^{u'}_{sh})/\tau}) \\
    &S^- = \sum_{u\neq u'}\exp({sim(F^u_{sp}, F^{u'}_{sp})/\tau}) + \sum_{u,u'}\exp({sim(F^u_{sp}, F^{u'}_{sh})/\tau})
\end{aligned}
\end{equation}
where $\tau$ serves as a temperature hyper-parameter for altering the strength of penalties on hard samples due to the modality gap \cite{wang2021understanding}. Then we take InfoNCE \cite{van2018representation} form function to compute the intra-sample contrastive learning loss $\mathcal L_{intraCL}$, which is formulated as:
\begin{equation}
\mathcal L_{intraCL} = -\mathbb{E}_{(F_{sp},F_{sh})\sim \mathcal{B}} \log \frac{S^+}{S^++S^-}
\end{equation}

Simultaneously, the inter-sample contrastive learning is adopted for the multimodal representations $F_M$ among diverse samples under the supervision of multimodal ground truth labels to further excavate the affective information inspired by SupCon \cite{khosla2020supervised}. Given a mini-batch of $\mathcal{B}=\{F^i_M\}^B_{i=1}$, we divide the representations into positive and negative sets according to the labels annotated as sentiment scores or emotion classes. For sentiment analysis task, we categorize the representations based on sentiment classes with a label threshold which decide the class each sentiment scores belongs to. While for emotion recognition and detection classes, we treat the representations with the same class as the positive pairs while the other representations as the negative pairs. Note that such setting is suitable for multi-label emotion recognition dataset \cite{zadeh2018mosei}, where we treat the representations with non-empty intersection set of emotion annotations as the positive pairs. To make the representations from various classes more discriminative with the guidance of multimodal labels, denoting positive pairs sets as $P\in\{F^{j}_M\}$, the inter-sample contrastive learning loss $\mathcal L_{interCL} $ is computed as:
\begin{equation}
\begin{aligned}
   \mathcal L_{interCL} = -\mathbb{E}_{F_M\sim \mathcal{B}} \log \frac{\sum^{j\neq j'}_{j,j'\in P}\exp({sim(F^{j}_M, F^{j'}_M)/\upsilon})}{\sum_{j,q\in B}\exp({sim(F^{j}_M, F^{q}_M)/\upsilon})}
\end{aligned}
\end{equation}
where $\upsilon$ is another temperature hyper-parameter to regulate the probability distribution over diverse instance samples \cite{hinton2015distilling}. 

Combining the intra- and inter-sample contrastive learning losses, the final semantic-centric contrastive learning loss $\mathcal L_{CL}$ is computed as:
\begin{equation}
\mathcal L_{CL} = \alpha\mathcal L_{intraCL} + \beta\mathcal L_{interCL} 
\end{equation}
where $\alpha$ and $\beta$ are hyper-parameters to adjust the contribution of each loss in the semantic-centric contrastive learning.

\subsection{Optimization Objective}
Regarding the learning of multimodal and semantic-centric representations as multi-task training paradigm where sub-task $*\in\{M,S,T,A,V\}$, we utilize various multi-layer perceptron (MLPs) as the mutli-task predictors to output the corresponding affective predictions, formulated as:
\begin{equation}
\begin{aligned}
   \hat{y}_M &= MLP(F_M;\theta_M) \in\mathbb{R}^r \\
   \hat{y}_* = MLP(F_*;\theta_*) &\in\mathbb{R}^r, \ \ F_*\in\{F_{sh},F^t_{sp},F^a_{sp},F^v_{sp}\}
\end{aligned}
\end{equation}
where $r=1$ denotes the sentiment scores as for sentiment analysis, $r=class$ denotes the number of emotion classes for recognition and $r=2$ denotes the binary classification for detection task. 

Along with the guidance of multimodal ground truth labels $y_{gt}$ and the supervision of the generated semantic-centic pseudo labels $p_*$ for each sub-task, the task prediction loss is formulated as:
\begin{equation}
    \mathcal L^*_{task} = \left\{
    \begin{aligned}
        & L2Loss(y, \hat{y}_*) = \frac{1}{N}\sum_{i=1}^{N} (y^i-\hat{y}^i_*)^2 \\
        & CrossEntropy(y, \hat{y}_*) = - \frac{1}{N}\sum_{i=1}^{N} y^i \log \hat{y}^i_*
    \end{aligned}
    \right.
\end{equation}
where $y$ denotes the ground truth $y_{gt}$ for multimodal sub-task and pseudo labels $p_*$ for the sub-tasks of semantic-centric representations. For multimodal sentiment analysis task, we utilize $L2\ Loss$ for the regression of sentiment scores; for multimodal emotion recognition and multimodal humor and sarcasm detection tasks, we adopt $CrossEntropy\ Loss$ for the classification of emotion or binary classes.

Lastly, the total optimization objective is formulated as:
\begin{equation}
    \mathcal L_{total} = \mathcal L_{CL} + \sum_{*\in\{M,S,T,A,V\}}\mathcal L^*_{task}
\end{equation}

\section{Experiments}
\subsection{Tasks and Datasets}

\begin{table*}[htbp]
\caption{Details of datasets in different MAC tasks, including data splitting and hyper-parameters settings. Note that for learning rate, '/' denotes 'learning rate for language model/learning rate for acoustic and visual modalies'. 'LR Scheduler' denotes the constant or cosine annealing scheduler for all learning rate after the warmup stage.}\label{table_dataset_hyperparameter}
\centering
\scalebox{0.97}{
\begin{tabular}{ccccccccc}
\toprule[1pt]
  MAC Task & \multicolumn{3}{c}{Multimodal Sentiment Analysis} & \multicolumn{3}{c}{Multimodal Emotion Recognition} & \multicolumn{2}{c}{Multimodal Humor/Sarcasm Detection}\\ 
\cmidrule(r){2-4}\cmidrule(r){5-7}\cmidrule(r){8-9}
  Dataset & CMU-MOSI & CMU-MOSEI & CH-SIMS (v2) & CMU-MOSEI & IEMOCAP & MELD & UR-FUNNY & MUStARD\\ 
\midrule[1pt]
  Train & 1,284 & 16,326 & 1,368 (2,722) & 16,322 & 5,228 & 9,765 & 7,614 & 554\\
  Valid & 229 & 1,871 & 456 (647) & 1,871 & 519 & 1,102 & 980 & 68\\
  Test & 686 & 4,659 & 457 (1,034) & 4,659 & 1,622 & 2,524 & 994 & 68\\
\midrule[1pt]
  Language Model & \multicolumn{3}{c}{BERT\cite{devlin2019bert} / XLNet\cite{yang2019xlnet}} & \multicolumn{3}{c}{sBERT\cite{reimers2019sbert}} & \multicolumn{2}{c}{BERT\cite{devlin2019bert} / RoBERTa\cite{liu2019roberta}} \\ 
  \cmidrule(r){2-4}\cmidrule(r){5-7}\cmidrule(r){8-9}
  Training epochs & 20 & 10 & 20 & 10 & 10 & 20 & 10 & 20 \\
  Batch size & 32 & 128 & 64 & 64 & 64 & 64 & 64 & 64 \\
  Learning rate & 4e-5/5e-4 & 4e-5/1e-4 & 3e-6/1e-4 & 5e-6/1e-4 & 1e-5/1e-4 & 1e-5/1e-4 & 1e-5/1e-4 & 5e-5/1e-4\\
  Weight decay & 5e-4 & 1e-3 & 1e-3 & 1e-2 & 1e-2 & 1e-3 & 1e-4 & 1e-3 \\
  LR Scheduler & Constant & Cosine & Constant & Constant & Constant & Cosine & Constant & Cosine \\
  Dropout & 0.1 & 0.1 & 0.1 & 0.3 & 0.2 & 0.2 & 0.3 & 0.1 \\
\bottomrule[1pt]
\end{tabular}
}
\end{table*}

\subsubsection{Multimodal sentiment analysis}
\textbf{CMU-MOSI} \cite{zadeh2016mosi} contains 2,199 monologue utterances clipped from 93 opinion videos spoken by 89 YouTube movie reviewers, which is annotated with a continuous sentiment score from $-3$ (strongly negative) to $+3$ (strongly positive).

\textbf{CMU-MOSEI} \cite{zadeh2018mosei} expands the size of dataset into 20k video clips segmented from 3,228 videos with 250 diverse topics collected by 1,000 distinct YouTube speakers, each of which is annotated for the sentiment on a $[-3, +3]$ Likert scale.

\textbf{CH-SIMS} \cite{yu2020chsims} collects 2,281 segments from 60 videos in  different movies, TV serials, and variety shows with spontaneous expressions, various head poses, occlusions, and illuminations performed by 474 distinct speakers
in Chinese. While \textbf{CH-SIMS v2} \cite{liu2022chsims2} doubles the scale of dataset by introducing more supervised and unsupervised instances with the same annotation method, where we only utilize the supervised ones in our experiments for fair comparision.

\subsubsection{Multimodal emotion recognition}
\textbf{CMU-MOSEI} \cite{zadeh2018mosei} annotates the utterance of each video into multiple emotional labels from $\{happy, sad, angry, surprise, disgust, fear\}$ as the settings of Ekman emotion classes \cite{ekman1980facial}.

\textbf{IEMOCAP} \cite{busso2008iemocap} provides 12 hours videos with two-way dialogues performed by 10 actors annotated into 6 classes $\{happy, sad, neutral, angry, excited, frustrated\}$. 

\textbf{MELD} \cite{poria2019meld} consists of about 13K utterances from 1,433 multi-party conversations from the TV-series \textit{Friends}, categories the emotion classes into 7 universal classes $\{neutral, surprise, fear, sadness, joy, disgust, angry\}$

\subsubsection{Multimodal humor and sarcasm detection}
\textbf{UR-FUNNY} \cite{hasan2019urfunny} comprises nearly 10K TED talk videos across 417 topics given by 1,741 different speakers, providing target punchline and the preceding context with even number of humor and non-humor instances.

\textbf{MUStARD} \cite{castro2019mustard} incorporates 690 videos containing target utterance along with associated historical dialogue, which are collected from famous TV shows including \textit{Friends}, \textit{The Big Bang Theory}, \textit{The Golden Girls} and \textit{Sarcasmaholics}, manually annotated with balanced numbers for the sarcasm property. 

\subsection{Evaluation Metrics}
We use public metrics of regression, recognition and detection  task to evaluate the performance of the proposed SemanticMAC framework and conduct fair comparison with baselines: For regression, seven-class/five-class/three-class classification accuracy (Acc7/Acc5/Acc3) indicating the correct sentiment label predictions in the label range; binary classification accuracy (Acc2) and F1-score are calculated with settings of positive and negative; mean absolute error (MAE) computing the average absolute difference between the final prediction and ground truth labels; Pearson correlation (Corr) measuring the degree of prediction skew. For recognition and detection, weighted accuracy (w-Acc) and F1-score (w-F1) along with weighted precision (w-Precision) and recall (w-Recall) score are computed according to the relative frequency of individual class; besides, standard accuracy (s-Acc), negative-weighted accuracy (n-Acc) and binary F1-score (b-F1) are reported according to dataset properties \cite{dai2020modality,dai2021multimodal,wu2023progressive}.

\begin{table*}[htbp]
\caption{Performance comparison between SemanticMAC and baselines on CMU-MOSI and CMU-MOSEI datasets for multimodal sentiment analysis task. The baseline models are reproduced with BERT as the language model.}
\label{exp_mosi_mosei}
\centering
\begin{tabular}{c|ccccc|ccccc}
    \toprule[1.5pt]
    \multirow{2}{*}{Models} & \multicolumn{5}{c|}{CMU-MOSI} & \multicolumn{5}{c}{CMU-MOSEI}\\
    & Acc7$\uparrow$  & Acc2$\uparrow$ & F1$\uparrow$ & MAE$\downarrow$ & Corr$\uparrow$ & Acc7$\uparrow$  & Acc2$\uparrow$ & F1$\uparrow$ & MAE$\downarrow$ & Corr$\uparrow$\\
    \midrule[1.5pt]
    EF-LSTM \cite{williams2018recognizing} & 34.5 &  79.0  &   78.9  & 0.952 & 0.651 &  
    49.3  & 80.3  &  81.0  & 0.603  & 0.682 \\
    LF-DNN \cite{williams2018dnn} & 33.6 &  79.3  & 79.3  &  0.978  &  0.658  &  
    52.1 & 82.3  & 82.2  & 0.561  & 0.723  \\
    TFN \cite{zadeh2017tensor} & 33.7  &  80.2  &  80.1  & 0.925  & 0.662  &  
    52.2 & 82.6  &  82.3  & 0.570 & 0.716 \\
    LMF \cite{liu2018efficient} & 32.7  &  80.1  & 80.0  & 0.931  & 0.670  & 
    52.0 & 83.7  &  83.8  & 0.568 & 0.727 \\
    MFN \cite{zadeh2018memory} & 34.2 &  80.0  &  80.0 & 0.951 & 0.665  & 
    51.1 & 84.0  &  83.9  & 0.575 & 0.720 \\
    Graph-MFN \cite{zadeh2018mosei} & 34.4 &  80.2  &  80.1  &  0.939  &  0.656  &  
     51.9  & 84.0  & 83.8  & 0.569  & 0.725  \\
    MFM \cite{tsai2019learning} & 33.3 &  80.0  &  80.1 & 0.948 & 0.664  &  
    50.8 & 83.4  & 83.4  & 0.580 & 0.722 \\
    MulT \cite{tsai2019multimodal} & 35.0 &  80.5  &  80.5  & 0.918 & 0.685  &
    52.1 & 84.0  &  83.9  & 0.564 & 0.732 \\
    MISA \cite{hazarika2020misa} & 43.5 &  83.5  &  83.5  & 0.752 & 0.784 & 
    52.2 & 84.3  &  84.3  & 0.550  & 0.758 \\
    MAG-BERT \cite{rahman2020integrating} & 45.1  &  84.6  &  84.6  &  0.730 & 0.789 &
    52.8 &  85.1  &  85.1  & 0.558  & 0.761 \\
    Self-MM \cite{yu2021learning} &  45.8 &  84.9  &  84.8  & 0.731  & 0.785  & 
    53.0 &  85.2  &  85.2  &  0.540 & 0.763 \\
    MMIM \cite{han2021improving} & 45.0 &  85.1  &  85.0  & 0.738  & 0.781  &  53.1 &  85.1  &  85.0  & 0.547  &  0.752 \\
    MMCL \cite{lin2022mmcl} & 46.5  &  86.3  &  86.2  &  0.705 & 0.797  & 53.6  &  85.9  &  85.7  & 0.537  &  0.765 \\
    MTMD \cite{lin2023mtmd} & 47.5  &  86.0  &  86.0  &  0.705 & 0.799  & 53.7  &  86.1  &  85.9  & 0.531  &  0.767 \\
    MissModal \cite{lin2023missmodal} & 47.2 & 86.1 & 86.0 & 0.698 & 0.801 & 53.9 & 85.9 & 85.8 & 0.533 & 0.769 \\
    \midrule[0.5pt] 
    SemanticMAC-BERT  & \textbf{48.3}  &   \textbf{86.4}  &   \textbf{86.4}  &  \textbf{0.685} & \textbf{0.811} & \textbf{54.5}  &  \textbf{87.3}  &  \textbf{87.2}  & \textbf{0.518}  &  \textbf{0.792} \\
    SemanticMAC-XLNet  & \textbf{49.3}  &  \textbf{88.0}  &  \textbf{88.0}  &  \textbf{0.632} & \textbf{0.845}  & \textbf{55.8}  &  \textbf{88.0}  &  \textbf{88.0}  & \textbf{0.497}  &  \textbf{0.807} \\
    \bottomrule[1.5pt]
\end{tabular}
\end{table*} 

\begin{table*}[htbp]
\caption{Performance comparison between SemanticMAC and baselines on CH-SIMS and CH-SIMSv2 datasets for multimodal sentiment analysis task. The baseline models are reproduced with BERT as the language model.}
\label{exp_sims}
\centering
\begin{tabular}{c|cccccc|cccccc}
    \toprule[1.5pt]
    \multirow{2}{*}{Models} & \multicolumn{6}{c|}{CH-SIMS} & \multicolumn{6}{c}{CH-SIMS v2}\\
    & Acc5$\uparrow$  & Acc3$\uparrow$ & Acc2$\uparrow$ & F1$\uparrow$ & MAE$\downarrow$ & Corr$\uparrow$ & Acc5$\uparrow$  & Acc3$\uparrow$ & Acc2$\uparrow$ & F1$\uparrow$ & MAE$\downarrow$ & Corr$\uparrow$\\
    \midrule[1.5pt]
    EF-LSTM \cite{williams2018recognizing} & 21.2 & 54.3 & 69.4 & 56.8 & 0.590 & 0.006 & 53.7 & 73.5 & 80.1 & 80.0 & 0.309 & 0.700 \\
    LF-DNN \cite{williams2018dnn} & 39.8 & 63.7 & 77.9 & 78.3 & 0.444 & 0.566 & 51.8 & 71.2 & 77.8 & 77.9 & 0.322 & 0.668 \\
    TFN \cite{zadeh2017tensor} & 40.9 & 65.8 & 77.6 & 77.6 & 0.429 & 0.587 & 53.3 & 70.9 & 78.1 & 78.1 & 0.322 & 0.662 \\
    LMF \cite{liu2018efficient} & 40.4 & 65.9 & 77.2 & 77.3 & 0.443 & 0.568 & 51.6 & 70.0 & 77.8 & 77.8 & 0.327 & 0.651 \\
    MFN \cite{zadeh2018memory} & 40.1 & 65.4 & 77.5 & 78.0 & 0.447 & 0.557 & 55.4 & 72.7 & 79.4 & 79.4 & 0.301 & 0.712 \\
    Graph-MFN \cite{zadeh2018mosei} & 41.9 & 66.5 & 78.2 & 78.4 & 0.438 & 0.579 & 48.9 & 68.6 & 76.6 & 76.6 & 0.334 & 0.644 \\
    MulT \cite{tsai2019multimodal} & 39.9 & 64.5 & 76.6 & 76.5 & 0.440 & 0.569 & 54.6 & 74.2 & 80.8 & 80.7 & 0.300 & 0.738 \\
    MISA \cite{hazarika2020misa} & 36.9 & 63.3 & 78.1 & 78.4 & 0.442 & 0.574 & 47.5 & 68.9 & 78.2 & 78.3 & 0.342 & 0.671 \\
    MAG-BERT \cite{rahman2020integrating} & 41.5 & 64.8 & 76.4 & 76.2 & 0.435 & 0.584 & 49.2 & 70.6 & 77.1 & 77.1 & 0.346 & 0.641 \\
    Self-MM \cite{yu2021learning} & 43.8 & 66.1 & 79.3 & 79.4 & 0.416 & 0.600 & 53.5 & 72.7 & 78.7 & 78.6 & 0.315 & 0.691 \\
    MMIM \cite{han2021improving} & 43.3 & 66.8 & 78.4 & 78.1 & 0.431 & 0.587 & 50.5 & 70.4 & 77.8 & 77.8 & 0.339 & 0.641 \\
    AV-MC \cite{liu2022chsims2} & 45.5 & 68.5 & 79.7 & 80.2 & 0.372 & 0.685 & 52.1 & 73.2 & 80.6 & 80.7 & 0.301 & 0.721\\
    \midrule[0.5pt] 
    SemanticMAC  & \textbf{47.2} & \textbf{72.5} & \textbf{84.8} & \textbf{84.8} & \textbf{0.366} & \textbf{0.718} & \textbf{55.3} & \textbf{75.1} & \textbf{83.8} & \textbf{83.7} & \textbf{0.293} & \textbf{0.771}\\
    \bottomrule[1.5pt]
\end{tabular}
\vspace{-0.2cm}
\end{table*} 

\subsection{Implementation Details}
All experiments are conducted on a single A100 GPU with CUDA 11.8. For each dataset, we convert the raw video into LMDB database for higher access speed in the end-to-end training and inference stage as Lei \textit{et al.} \cite{lei2021less}. Note that for fair comparison with baselines, we remain the same language model with the state-of-the-art models for each MAC task. Following Gkoumas \textit{et al.} \cite{gkoumas2021makes}, we present fifty-times random grid search to find the best hyper-parameters and we report the average results of 5 runs as the final performance. The splits of dataset and the settings of hyper-parameters are shown in Table \ref{table_dataset_hyperparameter}. We adopt AdamW \cite{loshchilov2018decoupled} as the optimizer and utilize a warmup strategy for all learning rates at the first epoch. For regression task, we utilize the minimum loss of validation set in the training stage as the reference to get the best parameters, while for recognition and detection tasks, we utilize w-F1 score of validation set as the one to determine the best model due to the confidence calibration issue \cite{guo2017calibration}.

\subsection{Baselines}
We report the results of baseline models by reproducing the corresponding open-source codes without extra mention. The baseline models are broadly categorized into: (1) Early and late fusion: \textbf{EF-LSTM}, \textbf{LF-LSTM}, \textbf{LF-Transformer}; (2) Tensor-based fusion models: \textbf{TFN} \cite{zadeh2017tensor}, \textbf{LMF} \cite{liu2018efficient}; (3) Explicitly intra- and inter-modal dynamics manipulation models: \textbf{MFN} \cite{zadeh2018memory}, \textbf{MFM} \cite{tsai2019learning}, \textbf{C-MFN} \cite{hasan2019urfunny}, \textbf{EmoEmbs} \cite{dai2020modality}; (4) Attention-based fusion models: \textbf{MulT} \cite{tsai2019multimodal}, \textbf{MISA} \cite{hazarika2020misa}, \textbf{MAG-BERT/MAG-XLNet} \cite{rahman2020integrating}, \textbf{TBJE} \cite{delbrouck2020btje}, \textbf{FE2E/MESM} \cite{dai2021multimodal}, \textbf{ME2ET} \cite{wu2023end}, \textbf{I-Attention} \cite{chauhan2020sentiment}, \textbf{BBFN} \cite{han2021bi}, \textbf{MuLoT} \cite{pramanick2022multimodal}; (5) Knowledge guidance models: \textbf{Self-MM} \cite{yu2021learning}, \textbf{MTMD} \cite{lin2023mtmd};  (6) Contrastive learning based models: \textbf{MMIM} \cite{han2021improving}, \textbf{MMCL} \cite{lin2022mmcl}; (7) Graph neural network based models: \textbf{Graph-MFN} \cite{zadeh2018mosei}, \textbf{DialogueGCN} \cite{ghosal2019dialoguegcn}, \textbf{MMGCN} \cite{hu2021mmgcn}, \textbf{COGMEN} \cite{joshi2022cogmen}, \textbf{CORECT} \cite{nguyen2023conversation}; (8) Context aware models: \textbf{bc-LSTM} \cite{poria2017context}, \textbf{CMN} \cite{hazarika2018conversational}, \textbf{ICON} \cite{hazarika2018icon}, \textbf{DialogueRNN} \cite{majumder2019dialoguernn}, \textbf{DialogueCRN} \cite{hu2021dialoguecrn},  \textbf{Multilogue-Net} \cite{shenoy2020multilogue}; (9) Data augmentation models: \textbf{AV-MC} \cite{liu2022chsims2}, \textbf{MissModal} \cite{lin2023missmodal}.

\begin{table*}[htbp]
\caption{Performance comparison between SemanticMAC and baselines on CMU-MOSEI dataset for multimodal emotion recognition task in the utterance scenario. $\dagger$ indicates the results copied from \cite{dai2021multimodal} and \cite{wu2023end}. }
\label{exp_mosei_emo_uttr}
\centering
\begin{tabular}{c|cccccccccccc|cc}
    \toprule[1.5pt]
    \multirow{2}{*}{Models} & \multicolumn{2}{c}{Happy} & \multicolumn{2}{c}{Sad} & \multicolumn{2}{c}{Anger} & \multicolumn{2}{c}{Surprise} & \multicolumn{2}{c}{Disgust} & \multicolumn{2}{c|}{Fear} & \multicolumn{2}{c}{Average} \\
    & n-Acc & b-F1 & n-Acc & b-F1 & n-Acc & b-F1 & n-Acc & b-F1 & n-Acc & b-F1 & n-Acc & b-F1 & n-Acc$\uparrow$ & b-F1$\uparrow$\\
    \midrule[1.5pt]
    LF-LSTM$^\dagger$ & 61.3 & 73.2 & 63.4 & 47.2 & 64.5 & 47.1 & 57.1 & 20.6 & 70.5 & 49.8 & 61.7 & 22.2 & 63.1 & 43.3 \\
    LF-Transformer$^\dagger$ & 60.6 & 72.9 & 60.1 & 45.5 & 65.3 & 47.7 & 62.1 & 24.2 & 74.4 & 51.9 & 62.1 & 24.0 & 64.1 & 44.4 \\
    EmoEmbs$^\dagger$ \cite{dai2020modality} & 61.2 & 71.9 & 60.5 & 47.5 & 66.8 & 49.4 & 63.3 & 24.0 & 69.6 & 48.7 & 63.8 & 23.4 & 64.2 & 44.2 \\
    MulT$^\dagger$ \cite{tsai2019multimodal} & 67.2 & 75.4 & 64.0 & 48.3 & 64.9 & 47.5 & 61.4 & 25.6 & 71.6 & 49.3 & 62.9 & 25.3 & 65.4 & 45.2 \\
    FE2E$^\dagger$ \cite{dai2021multimodal} & 65.4 & 72.6 & 65.2 & 49.0 & 67.0 & 49.6 & 66.7 & \textbf{29.1} & 77.7 & \textbf{57.1} & 63.8 & 26.8 & 67.6 & 47.4 \\
    MESM$^\dagger$ \cite{dai2021multimodal} & 64.1 & 72.3 & 63.0 & 46.6 & 66.8 & 49.3 & 65.7 & 27.2 & 75.6 & 56.4 & 65.8 & 28.9 & 66.8 & 46.8 \\
    ME2ET$^\dagger$ \cite{wu2023end} & 66.4 & 73.2 & 66.2 & 50.0 & 67.9 & 51.1 & 63.3 & 27.7 & 76.4 & 56.4 & 69.3 & 29.3 & 68.3 & 48.0 \\
    \midrule[0.5pt]
    SemanticMAC  & \textbf{73.2} & \textbf{75.2} & \textbf{69.1} & \textbf{51.7} & \textbf{69.7} & \textbf{52.6} & \textbf{66.9} & 26.7 & \textbf{76.7} & 54.3 & \textbf{71.1} & \textbf{30.2} & \textbf{71.0} & \textbf{48.5}\\
    \bottomrule[1.5pt]
\end{tabular}
\end{table*} 
\begin{table*}[htbp]
\caption{Performance comparison between SemanticMAC and baselines on IEMOCAP dataset for multimodal emotion recognition task in the utterance scenario. $\dagger$ indicates the results copied from \cite{dai2021multimodal} and \cite{wu2023end}. }
\label{exp_iemocap_emo_uttr}
\centering
\begin{tabular}{c|cccccccccccc|cc}
    \toprule[1.5pt]
    \multirow{2}{*}{Models} & \multicolumn{2}{c}{Happy} & \multicolumn{2}{c}{Sad} & \multicolumn{2}{c}{Neutral} & \multicolumn{2}{c}{Anrgy} & \multicolumn{2}{c}{Excited} & \multicolumn{2}{c|}{Frustrated} & \multicolumn{2}{c}{Average} \\
    & s-Acc & b-F1 & s-Acc & b-F1 & s-Acc & b-F1 & s-Acc & b-F1 & s-Acc & b-F1 & s-Acc & b-F1 & s-Acc$\uparrow$ & b-F1$\uparrow$\\
    \midrule[1.5pt]
    LF-LSTM$^\dagger$ & 67.2 & 37.6 & 78.2 & 54.0 & 66.5 & 47.0 & 71.2 & 49.4 & 79.3 & 57.2 & 68.2 & 51.5 & 71.8 & 49.5 \\
    LF-Transformer$^\dagger$ & 85.2 & 37.6 & 87.4 & 57.4 & 72.4 & 49.7 & 81.9 & 50.7 & 85.3 & 57.3 & 60.5 & 49.3 & 78.8 & 50.3 \\
    EmoEmbs$^\dagger$ \cite{dai2020modality} & 69.6 & 38.3 & 80.8 & 53.0 & 73.6 & 48.7 & 65.9 & 48.9 & 73.5 & 58.3 & 68.5 & 52.0 & 72.0 & 49.8 \\
    MulT$^\dagger$ \cite{tsai2019multimodal} & 80.0 & 46.8 & 83.5 & 65.4 & 74.9 & 53.7 & 77.9 & 60.7 & 76.9 & 58.0 & 72.4 & 57.0 & 77.6 & 56.9 \\
    FE2E$^\dagger$ \cite{dai2021multimodal} & 90.0 & 44.8 & 89.1 & 65.7 & 79.1 & 58.4 & 88.7 & 63.9 & 89.1 & 61.9 & 71.2 & 57.8 & 84.5 & 58.8 \\
    MESM$^\dagger$ \cite{dai2021multimodal} & 89.5 & 47.3 & 88.6 & 62.2 & 77.0 & 52.0 & 88.2 & 62.8 & 88.3 & 61.2 & 74.9 & 58.4 & 84.4 & 57.4 \\
    ME2ET$^\dagger$ \cite{wu2023end} & 90.0 & 44.7 & 92.4 & 73.8 & 78.8 & 58.7 & 89.8 & 65.9 & 89.2 & 63.9 & 79.2 & 60.7 & 86.5 & 61.3 \\
    \midrule[0.5pt]
    SemanticMAC  & \textbf{92.3} & \textbf{56.6} & \textbf{94.2} & \textbf{79.9} & \textbf{82.7} & \textbf{61.1} & \textbf{90.8} & \textbf{67.8} & \textbf{92.3} & \textbf{74.2} & \textbf{79.8} & \textbf{63.8} & \textbf{88.7} & \textbf{67.2}\\
    \bottomrule[1.5pt]
\end{tabular}
\end{table*} 

\subsection{Experiment Results}
For \textit{multimodal sentiment analysis task}, as shown in Table \ref{exp_mosi_mosei}, SemanticMAC reaches the state-of-the-art performance when utilizing BERT as the same language model as the baselines. Besides, comparing the improvement over the best baseline models on CMU-MOSI and CMU-MOSEI datasets, higher performance gains can be obtained by training SemanticMAC with a larger scale of dataset. Moreover, the proposed architecture can be further generalized to other language model such as XLNet without modifications, which achieves more superior performance on all metrics. Additionally, as shown in Table \ref{exp_sims}, since CH-SIMS (v2) dataset is collected in Chinese environment, the results demonstrate that the visual and audio features extracted by ImageBind and learned by the proposed Affective Perceiver can be effectively utilized to boost the performance of language model in multilingual settings. We reckon that this is mainly attributed to the fact that the affective information contained in vision and audio data is mostly independent with specific language.

\begin{table}[htbp]
\scriptsize
\setlength{\tabcolsep}{5.5pt}
\caption{Performance comparison between SemanticMAC and baselines on CMU-MOSEI for multimodal emotion recognition task in the conversation scenario. $\dagger$ indicates results from \cite{nguyen2023conversation}.}
\label{exp_mosei_emo_conv}
\centering
\begin{tabular}{c|cccccc} 
    \toprule[1.5pt]
    \multirow{2}{*}{Models} & \multicolumn{6}{c}{w-F1 of emotions in CMU-MOSEI $\uparrow$} \\ 
     & Happy & Sad & Anger & Surprise & Disgust & Fear \\ 
    \midrule[1.5pt]
    Multilogue-Net$^\dagger$ \cite{shenoy2020multilogue} & 67.8 & 65.3 & 67.0 & 86.1 & 74.9 & 87.8 \\
    TBJE$^\dagger$ \cite{delbrouck2020btje} & 65.9 & 70.8 & 70.9 & 86.0 & 82.6 & 87.8 \\
    COGMEN$^\dagger$ \cite{joshi2022cogmen} & 70.9 & 70.9 & 74.2 & 86.5 & 84.3 & 87.8 \\ 
    CORECT$^\dagger$ \cite{nguyen2023conversation} & 71.4 & 72.9 & 76.8 & 86.5 & 84.3 & 87.9 \\
    \midrule[0.5pt]
    SemanticMAC  & \textbf{71.9} & \textbf{73.8} & \textbf{77.4} & \textbf{86.6} & \textbf{84.6} & \textbf{88.5}\\
    \bottomrule[1.5pt]
\end{tabular}
\end{table} 

For \textit{multimodal emotion recognition task}, as shown in Table \ref{exp_mosei_emo_uttr} -\ref{exp_iemocap_emo_meld_conv}, we compare SemanticMAC with the baselines in both conversation and utterance settings, where the former means feeding all context inside the dialogue into the models while the latter denotes utilizing the single target utterance as input. Note that on both specific emotion class and average accuracy, SemanticMAC mostly outperforms the baseline models no matter training on CMU-MOSEI, IEMOCAP or MELD datasets without any usage of speaker information. Meanwhile, SemanticMAC surpasses recent graph-based models \cite{ghosal2019dialoguegcn,joshi2022cogmen,nguyen2023conversation} and context-aware models \cite{hu2021dialoguecrn,majumder2019dialoguernn,shenoy2020multilogue} in needless of graph neural networks or delicate context-aware module which have been proved effective in constructing the complicated emotion relationships of utterances in conversation. The reason is that as the weak supervision in the training of unimodal representations, semantic-centric label succeeds in integrating emotion-related semantics and tackling the semantic mismatch among representations with various emotion classes. 

\begin{table*}[htbp]
\setlength{\tabcolsep}{3pt}
\caption{Performance comparison between SemanticMAC and baselines on IEMOCAP and MELD dataset for multimodal emotion recognition task in the conversation scenario. The weighted F1 (w-F1) of each emotion class is reported for fine-grained comparison. The baseline models are reproduced with the corresponding open-source codes.}
\label{exp_iemocap_emo_meld_conv}
\centering
\begin{tabular}{c|cccccc|cc|ccccc|cc}
    \toprule[1.5pt]
    \multirow{2}{*}{Models} & \multicolumn{6}{c|}{w-F1 of emotions in IEMOCAP $\uparrow$} & \multicolumn{2}{c|}{Average} & \multicolumn{5}{c|}{w-F1 of emotions in MELD $\uparrow$} & \multicolumn{2}{c}{Average}\\
    & Happy & Sad & Neutral & Angry & Excited & Frustrated & w-Acc$\uparrow$ & w-F1$\uparrow$ & Neutral & Surprise & Sadness & Joy & Anger & w-Acc$\uparrow$ & w-F1$\uparrow$\\
    \midrule[1.5pt]
    bc-LSTM \cite{poria2017context} & 32.4 & 73.0 & 54.3 & 63.4 & 60.9 & 61.6 & 59.4 & 59.1 & 76.1 & 47.4 & 21.4 & 53.1 & 40.4 & 59.5 & 56.9 \\
    CMN \cite{hazarika2018conversational} & 29.7 & 72.7 & 56.6 & 65.0 & 68.5 & 63.3 & 62.1 & 61.3 & - & - & - & - & - & - & - \\
    ICON \cite{hazarika2018icon} & 31.6 & 72.1 & 61.0 & 66.6 & 68.5 & 64.5 & 63.4 & 62.8 &  - & - & - & - & - & - & - \\
    DialogueRNN \cite{majumder2019dialoguernn} & 34.8 & 78.2 & 55.2 & 62.0 & 69.3 & 58.9 & 61.2 & 61.0 & 75.6 & 47.3 & 26.7 & 50.9 & 45.8 & 59.4 & 57.5 \\
    DialogueGCN \cite{ghosal2019dialoguegcn} & 41.9 & 78.0 & 58.7 & 56.1 & 73.6 & 58.3 & 63.3 & 62.5 & 75.3 & 47.5 & 17.4 & 50.5 & 39.2 & 58.1 & 55.7 \\
    MMGCN \cite{hu2021mmgcn} & 41.1 & 78.3 & 60.4 & \textbf{68.5} & 73.6 & 61.5 & 65.2 & 64.9 & 76.3 & 46.9 & 16.8 & 53.4 & 44.8 & 60.5 & 57.3 \\
    DialogueCRN \cite{hu2021dialoguecrn} & \textbf{60.1} & 82.0 & 59.8 & 62.9 & 75.8 & 58.5 & 66.4 & 66.3 & \textbf{81.8} & 46.3 & 16.8 & 57.1 & 40.0 & 59.9 & 57.3 \\
    COGMEN \cite{joshi2022cogmen} & 53.5 & 78.1 & 65.1 & 66.6 & 70.9 & 63.8 & 66.8 & 66.9 & 74.7 & 49.8 & 25.1 & 50.9 & 44.6 & 58.9 & 57.1 \\
    CORECT \cite{nguyen2023conversation} & 56.6 & 81.0 & 63.9 & 65.8 & 71.7 & 62.7 & 67.3 & 67.2 & 76.7 & 48.4 & 29.4 & 51.7 & 43.2 & 60.7 & 58.4 \\
    \midrule[0.5pt] 
    SemanticMAC  & 53.4 & \textbf{82.3} & \textbf{69.5} & 65.9 & \textbf{82.5} & \textbf{66.3} & \textbf{70.4} & \textbf{69.8} & 77.4 & \textbf{54.3} & \textbf{35.0} & \textbf{57.4} & \textbf{46.9} & \textbf{62.2} & \textbf{61.4}\\
    \bottomrule[1.5pt]
\end{tabular}
\end{table*} 

For \textit{multimodal humor and sarcasm detection} task, as shown in Table \ref{exp_urfunny_mustard}, compared with previous models relying on manual extracted features provided by the original datasets, SemanticMAC detects the intention of the target utterance more accurately for humor and sarcasm with BERT on UR-FUNNY and MUStARD datasets. When training with RoBERTa, SemanticMAC reaches higher performance and defeats the baseline with larger language model \cite{hyun2024smile}. The advanced results indicate the efficiency of SemanticMAC in capturing the contradictory correlation among the punchline and context to predict the humorous and sarcastic anchors.

\begin{table*}[htbp]
\caption{Performance comparison between SemanticMAC and baselines on UR-FUNNY and MUStARD dataset for multimodal humor and sarcasm detection task. $\dagger$ indicates the results copied from \cite{pramanick2022multimodal} or the original papers.}
\label{exp_urfunny_mustard}
\centering
\begin{tabular}{c|cccc|cccc}
    \toprule[1.5pt]
    \multirow{2}{*}{Models} & \multicolumn{4}{c|}{UR-FUNNY} & \multicolumn{4}{c}{MUStARD} \\
    & w-Precision $\uparrow$ & w-Recall $\uparrow$ & w-Acc $\uparrow$ & w-F1 $\uparrow$ & w-Precision $\uparrow$ & w-Recall $\uparrow$ & w-Acc $\uparrow$ & w-F1 $\uparrow$\\
    \midrule[1.5pt]
    C-MFN$^\dagger$ \cite{hasan2019urfunny} & - & - & 65.23 & - & - & - & 70.00 & - \\
    SVM$^\dagger$ \cite{cortes1995svm} & - & - & - & - & 72.00 & 71.60 & - & 71.60 \\
    I-Attention$^\dagger$ \cite{chauhan2020sentiment} & - & - & - & - & 73.40 & 72.75 & - & 72.57 \\
    TFN$^\dagger$ \cite{zadeh2017tensor} & - & - & 64.71 & - & - & - & 68.57 & - \\
    MISA$^\dagger$ \cite{hazarika2020misa}& 71.62 & 70.61 & 70.61 & 69.82 & - & - & 66.18 & - \\
    BBFN$^\dagger$ \cite{han2021bi}& 71.96 & 71.68 & 71.68 & 71.36 & - & - & 71.42 & - \\
    MAG-XLNet$^\dagger$ \cite{rahman2020integrating}& - & - & 72.43 & - & - & - & 74.72 & - \\
    MuLoT$^\dagger$ \cite{pramanick2022multimodal}& - & - & 73.97 & - & - & - & 76.82 & - \\
    SMILE-LLaMA$^\dagger$ \cite{hyun2024smile}& - & - & 75.10 & - & - & - & 77.50 & - \\
    \midrule[0.5pt]
    SemanticMAC-BERT  & \textbf{74.89} & \textbf{74.48} & \textbf{74.48} & \textbf{74.40} & \textbf{80.52} & \textbf{79.69} & \textbf{79.69} & \textbf{79.61}\\
    SemanticMAC-RoBERTa  & \textbf{76.08} & \textbf{75.60} & \textbf{75.60} & \textbf{75.53} & \textbf{84.76} & \textbf{81.25} & \textbf{81.25} & \textbf{80.88}\\
    \bottomrule[1.5pt]
\end{tabular}
\vspace{-0.2cm}
\end{table*} 

\subsection{Ablation Study}

\begin{table}[htbp]
\caption{Ablation study of SemanticMAC with BERT as the language model on CMU-MOSEI dataset.}
\label{ablation_mosei}
\centering
\setlength\tabcolsep{5.1pt}
\begin{tabular}{l|ccccc}
    \toprule[1.5pt]
    \multicolumn{1}{c|}{Description} & Acc7$\uparrow$  & Acc2$\uparrow$ & F1$\uparrow$ & MAE$\downarrow$ & Corr$\uparrow$ \\
    \midrule[1.5pt]
    \multicolumn{1}{c|}{SemanticMAC} & \textbf{54.5}  &  \textbf{87.3}  &  \textbf{87.2}  & 0.518  &  \textbf{0.792} \\
    \midrule[0.5pt]
    \multicolumn{6}{l}{(1) Affective Perceiver} \\
    w/o $E_{fr}$ & 53.9 & 86.9 & 86.9 & 0.521 & 0.787  \\
    w/o layer normalization & 53.8 & 86.9 & 86.8 & \textbf{0.517} & 0.788  \\
    w/o residual connection & 53.6 & 87.0 & 86.9 & 0.522 & 0.787  \\
    \midrule[0.5pt]
    \multicolumn{6}{l}{(2) Semantic-centric Gated Feature Interaction (SGFI)} \\
    w/o $E_{md}$ & 54.0 & 87.0 & 87.0 & 0.521 & 0.787  \\
    w/o multi-query attention & 53.7 & 86.8 & 86.8 & 0.524 & 0.786  \\
    w/o Bridge Tokens $B$ & 54.0 & 86.9 & 86.9 & 0.521 & 0.787  \\
    w/o gated ReLU & 53.5 & 86.9 & 86.8 & 0.523 & 0.785  \\
    \midrule[0.5pt]
    \multicolumn{6}{l}{(3) Semantic-centric Label Generation (SCLG)} \\
    w/o momentum updating & 53.5 & 86.8 & 86.7 & 0.527 & 0.781 \\
    rp pseudo label $p_*$ with $y_{gt}$ & 53.3 & 86.2 & 86.1 & 0.529 & 0.780 \\
    w/o multi-task (only $\mathcal L^M_{task}$) & 53.0 & 86.3 & 86.2 & 0.532 & 0.775 \\
    \midrule[0.5pt]
    \multicolumn{6}{l}{(4) Semantic-centric Contrastive Learning (SCCL)} \\
    w/o $\mathcal L_{intraCL}$ & 53.1 & 86.3 & 86.2 & 0.531 & 0.780 \\
    w/o $\mathcal L_{interCL}$ & 53.6 & 86.6 & 86.6 & 0.524 & 0.785 \\
    w/o $\mathcal L_{CL}$ & 53.0 & 86.4 & 86.5 & 0.537 & 0.784 \\
    \bottomrule[1.5pt]     
\end{tabular}
\vspace{-0.4cm}
\end{table} 

To further reveal the contributions of different modules inside the proposed architecture, we perform ablation study for SemanticMAC on CMU-MOSEI dataset as shown in Table \ref {ablation_mosei}. Firstly, when capturing intra- and inter-modal dynamics in Affective Perceriver and SGFI, the learnable frame embeddings $E_{fr}$ and the modality embeddings $E_{md}$ are productive in assigning temporal information for both audio and vision modalities and revealing the type of modality in multimodal fusion. Moreover, multi-query attention and Bridge Tokens $B$ are effective in decreasing the modality gap and exploiting the common semantics when conducting cross-modal attention in feature interaction. The gated ReLU succeeds in filtering unrelated noise, leading to performance decrease on Acc7 lacking of the gated mechanism. 

Additionally, SemanticMAC adopts semantic-centric labels $p_*$ in SCLG to guide the learning of multiple sub-tasks. Therefore, when replacing pseudo labels $p_*$ with the ground truth labels $y_{gt}$ in the learning of semantic-specific and semantic-shared representations $F^u_{sp}$ and $F_{sh}$, the model suffer from the issue of semantic mismatch, leading to huge performance drop on all metrics. Besides, the momentum updating strategy is capable of stabilizing the generation process of semantic-centric labels. Lastly, the intra- and inter-sample contrastive learning in SCCL are both beneficial of distinguishing representations according to semantics and multimodal ground truth, so that optimize without contrastive learning largely hurt the final performance.

\section{Further Analysis}
\begin{figure*}[htbp]
\centering
\vspace{-0.5cm}
 \subfigure[Feature Distribution]{
 \begin{minipage}[b]{0.21\linewidth}
    \centering
    \hspace{-0.5cm}
    \includegraphics[scale=0.43]{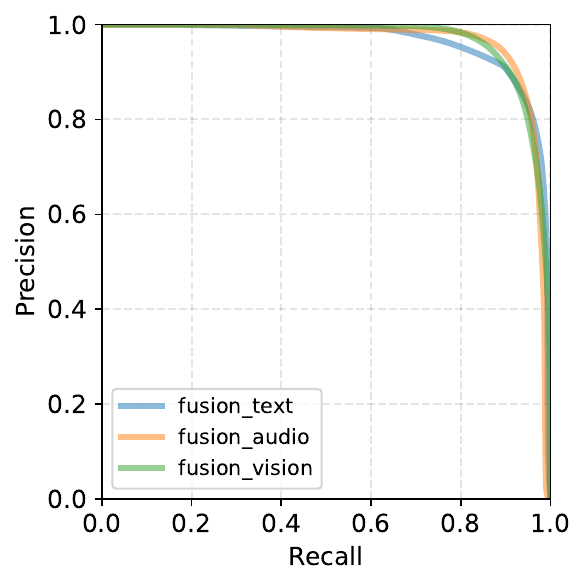} 
    \vspace{0.1cm}
    \\
    \hspace{-0.6cm}
    \includegraphics[scale=0.15]{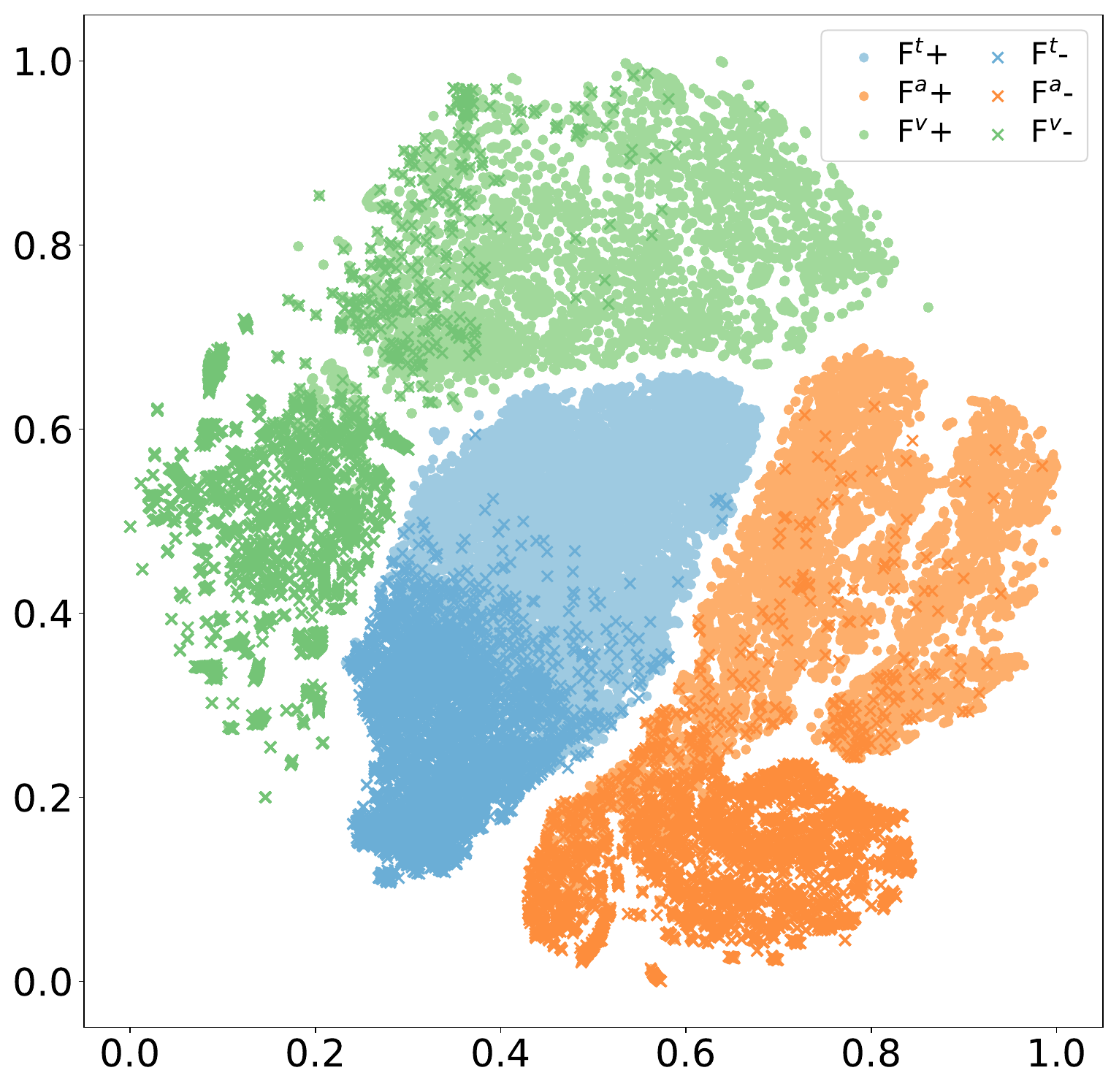} 
 \end{minipage}
 \label{vis_feature_distribution_msa}
 }
 \subfigure[Label Distribution on sentiment analysis]{
 \begin{minipage}[b]{0.33\linewidth}
		\includegraphics[scale=0.4]{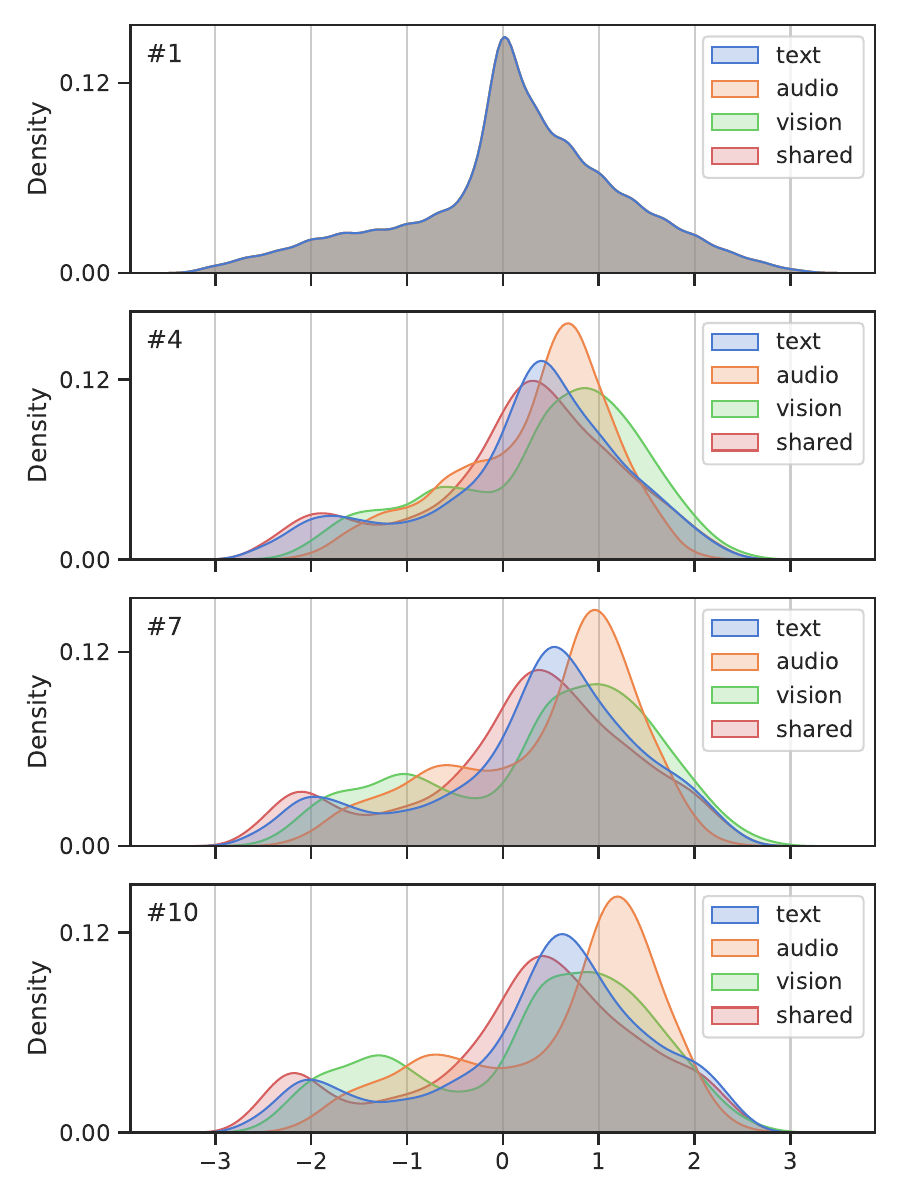}
  \label{vis_label_distribution_msa}
    \vspace{-0.3cm}
 \end{minipage}
 }
 \subfigure[Label Distribution on emotion recognition]{
 \begin{minipage}[b]{0.36\linewidth}
		\includegraphics[scale=0.4]{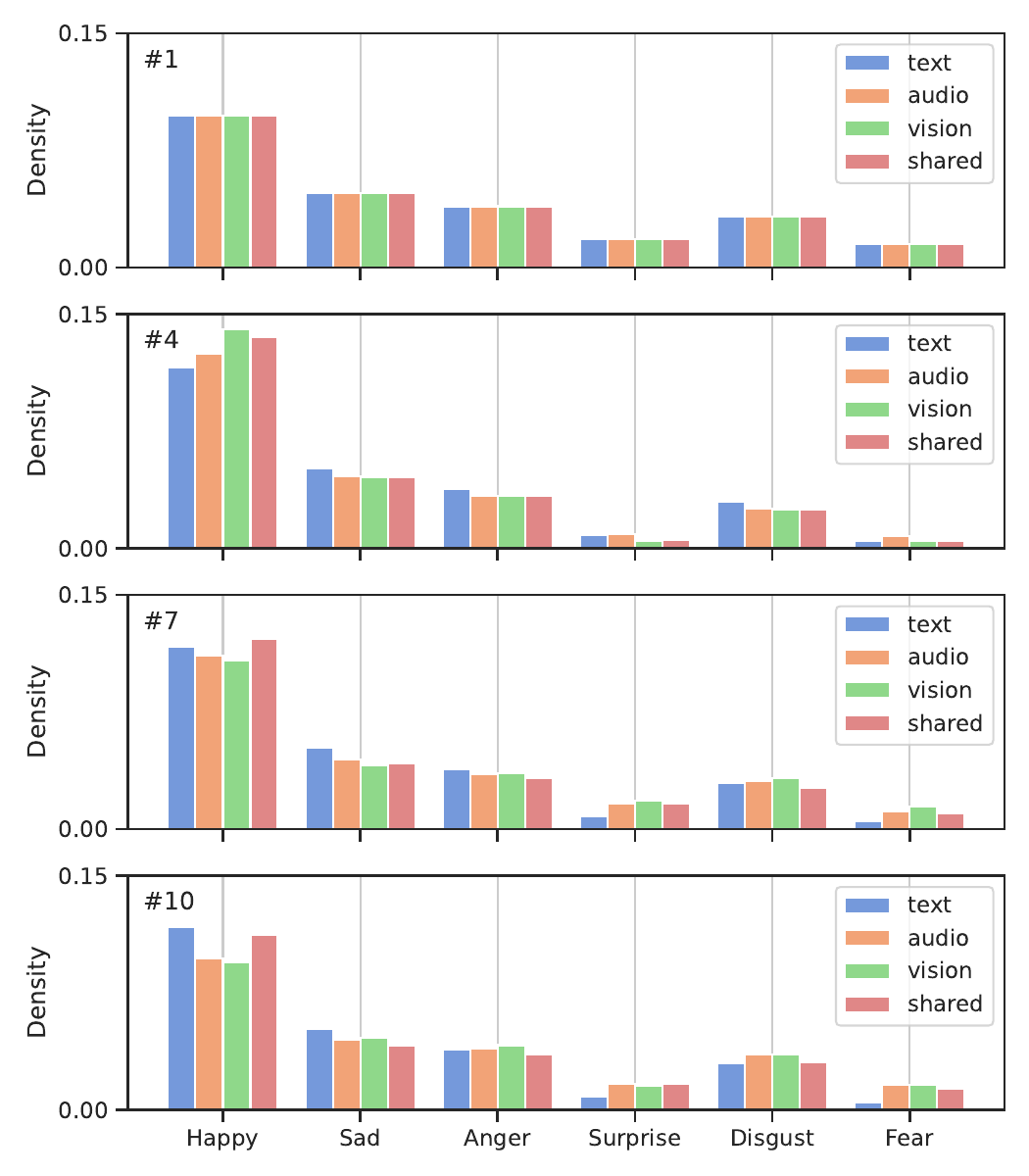}
  \label{vis_label_distribution_mer}
    \vspace{-0.3cm}
 \end{minipage}
 }
\caption{The distribution of semantic-centric features and labels for SemanticMAC on CMU-MOSEI dataset. (a) The upper left figure is the PR curve of the fusion multimodal representations and the unimodal representations training with BERT. The lower left one is the distribution of semantic-specific representations $F_t, F_a, F_v$ in the feature space where $+/-$ denotes the positive/negative sentiment intensity with pseudo labels. The right ones are the variations of semantic-centric labels distribution on (b) multimodal sentiment analysis task and (c) multimodal emotion recognition task with \#1,4,7,10 $epoch$. }\label{fig_visualization}
\vspace{-0.3cm}
\end{figure*}

\subsection{Effect in tackling semantic imbalance}
Aiming at addressing the issue of semantic imablance shown in Figure \ref{fig_PRcurve_sota}, we visualize the Precision-Recall curve of SemanticMAC with BERT on CMU-MOSEI to reveal the semantic abundance of various unimodal representations as shown in the Figure \ref{vis_feature_distribution_msa}. By replacing the manual acoustic and visual features with ImageBind in SemanticMAC, the contribution of unimodal features from different modalities are well balanced in the final multimodal representations. Besides, Affective Perceiver competently integrates the affective information into learnable unimodal tokens, increasing the semantic abundance of acoustic and visual representations. As the end-to-end training processes, the issue of semantic imbalance can be practically tackled in the stage of multimodal fusion, which further demonstrates the effectiveness of the proposed architecture. 

\subsection{Visualization in the Embedding Space}
To better reveal the distribution of the semantic-specific features, we utilize T-SNE \cite{hinton2008visualizing} to visualize representations trained on CMU-MOSEI in the embedding space. As shown in Figure \ref{vis_feature_distribution_msa}, for $u\in\{t,a,v\}$, the semantic-specific representations $F^u_{sp}$ are discriminative according to the semantic-centric labels, regardless of the observable modality gap. Meanwhile, the representations with consistent sentiment from various modalities are well classified into the opposite ends of the embedding space, indicating the productivity of semantic-centric cross-modal feature interaction.

\subsection{Distribution of Semantic-centric Labels}
We present case study to concretize the issue of semantic mismatch as shown in Table \ref{case_study_mismatch}. For sentiment analysis (\#1-\#3), the intensity of generated pseudo labels $p_*$ for various modalities are quite different, even contradictory due to the exclusive content contained in semantic-specific representations. The similar trend can be observed when classifying emotion class for unimodal and multimodal representations in the examples (\#4-\#6) of multimodal emotion recognition. Besides, the semantic-centric labels are discrepant from the value or class of ground truth labels $y_{gt}$, implicitly validating that the pseudo labels are productive in capturing the semantics information contained in diverse representations.

To further verify the effectiveness of semantic-centric label in tackling semantic mismatch, we visualize the generation process of semantic-centric labels during training on sub-tasks $*\in\{S,T,A,V\}$ in Figure \ref{fig_visualization}. Notes that the multimodal labels of the $1^{st}$ epoch denotes the ground truth labels. As the training stage proceeds, the distributions of semantic-centric labels for diverse semantic-specific and -shared representations vary in distinctive ways, demonstrating the pseudo labels can be generated according to different semantic-centric representations. For multimodal sentiment analysis in Figure \ref{vis_label_distribution_msa}, the pseudo labels polarize with more discriminative sentiment tendency, where more samples are assigned with positive or negative intensity. For multimodal emotion recognition in Figure \ref{vis_label_distribution_mer}, the frequency of emotion classes in semantic-centric labels are rearranged by the affective semantics in various modalities. Besides, for the emotion classes such as happy and sad which can be expressed explicitly in language, the pseudo labels are more frequently arranged for textual modality. While for the emotion more likely to be revealed in expressive face or tune such as surprise and fear, the pseudo labels are more presented in audio and vision modalities.

\subsection{Unimodal Feature Performance Comparison}
Aiming at validating the effectiveness of acoustic and visual features learned by ImageBind and Affective Perceiver, we conduct experiments on $X_u$ and $F_u$ in the architecture by linear probing \cite{he2022masked} for the sole audio and vision modality $u\in\{a,v\}$ on CMU-MOSEI. Note that the prediction of unimodal features is attained through conducting average pooling and linear projection trained on the learned and manual features with other parameters frozen. As shown in Table \ref{exp_unimodal}, compared with manually extracted features, both of unimodal embeddings $X_u$ and representations $F_u$ achieve superior performance on both regression and classification metrics. The performance of $X_u$ reveals the multimodal alignment and generalization power of ImageBind. Additionally, Affective Perceiver productively filters the noise of $X_u$ and integrates the affective information in $F_u$, leading to higher performance. 
\begin{table}[htbp]
\vspace{-0.2cm}
\caption{Performance comparison through linear probing of acoustic and visual features learned by SemanticMAC and extracted by commonly-used manual toolkit CMU-MultimodalSDK.}
\label{exp_unimodal}
\centering
\begin{tabular}{cc|ccccc}
    \toprule[1.5pt]
    \multicolumn{2}{c|}{Unimodal Feature}  & Acc3$\uparrow$ & Acc2$\uparrow$ & F1$\uparrow$ & MAE$\downarrow$ & Corr$\uparrow$ \\ 
    \midrule[1.5pt]
    \multirow{3}{*}{Audio} & Manual & 42.3 & 64.7 & 60.8 & 0.824 & 0.196 \\
    & $X_a$ & 46.1 & 71.4 & 69.8 & 0.798 & 0.331 \\
    & $F_a$ & \textbf{48.5} & \textbf{72.3} & \textbf{71.3} & \textbf{0.774} & \textbf{0.433} \\
    \midrule[0.5pt]
    \multirow{3}{*}{Vision} & Manual & 43.5 & 64.4 & 60.5 & 0.818 & 0.204 \\
    & $X_v$ & 44.9 & 73.9 & 72.9 & 0.790 & 0.351 \\
    & $F_v$ & \textbf{50.3} & \textbf{74.5} & \textbf{73.6} & \textbf{0.770} &  \textbf{0.450} \\
    \bottomrule[1.5pt]
\end{tabular}
\vspace{-0.4cm}
\end{table} 

\subsection{Influence of Fixed and Varying Video Length}
To demonstrate the effectiveness of SemanticMAC in processing videos with various length, we conduct experiments in the settings of both fixed and various frames (in both audio and vision streams) on CMU-MOSEI, which has a wide range of video lengths from $0.7\ s$ to $108.9\ s$ \cite{zadeh2018mosei}. As shown in \ref{exp_fixed_various_frame},
the model trained on videos with various frames outperform the ones trained on videos with fixed frames. Besides, either too few or too much frames are not beneficial for the information extraction of Affective Perceiver or the feature interaction among different modalities, remaining consistent trend with sparse to dense uniform sampling \cite{lei2021less}. This indicates the importance of balancing the information redundancy and semantic abundance for the performance of affective computing model. Therefore, the ability of handling videos with various length results in higher robustness and applicability when adopting SemanticMAC in diverse downstream scenarios.
\begin{table}[htbp]
\vspace{-0.2cm}
\caption{Performance comparison between the video data with the settings of fixed and various frames for SemanticMAC.}
\label{exp_fixed_various_frame}
\centering
\begin{tabular}{cc|ccccc}
    \toprule[1.5pt]
    \multicolumn{2}{c|}{Frame Setting} & Acc7$\uparrow$  & Acc2$\uparrow$ & F1$\uparrow$ & MAE$\downarrow$ & Corr$\uparrow$ \\
    \midrule[1.5pt]
    \multirow{2}{*}{\makecell{Fixed \\(frames/video)}} & 5 & 52.9 & 86.0 & 85.9 & 0.533 & 0.773 \\
    & 100 & 52.6 & 85.8 & 85.7 & 0.536 & 0.775 \\
    \midrule[0.5pt]
    \multicolumn{2}{c|}{Various frames} & \textbf{54.5}  &  \textbf{87.3}  &  \textbf{87.2}  & \textbf{0.518}  &  \textbf{0.792} \\
    \bottomrule[1.5pt]
\end{tabular}
\vspace{-0.4cm}
\end{table} 

\section{Conclusion}
In this paper, we proposed a novel end-to-end multimodal affective computing framework, SemanticMAC, to effectively learn semantic-specific and -shared representations with the supervision of the generated semantic-centric labels. Extensive experiments on 7 public video-based datasets in 4 downstream MAC tasks demonstrate the effectiveness of the proposed approach. The visualization and ablation study consistently reveals that SemanticMAC productively tackles the challenges of semantic imbalance and semantic mismatch for various modalities. 

In the future, we will utilize recent emerging large language models to promote higher performance of the proposed method, since SemnaticMAC has been verified universally across different language models. Moreover, we tend to extend the end-to-end pipeline for multimodal affective computing in more downstream applications of human-AI interaction.

\bibliographystyle{IEEEtran}
\bibliography{custom}

\vfill

\end{document}